# Post-Hoc Calibrated Anomaly Detection
## *Post-hoc kalibrierte Anomalieerkennung*

by

Sean Gloumeau

Submitted to the Fachbereich Informatik
in partial fulfillment of the requirements for the degree of

EUROPEAN MASTERS IN EMBEDDED COMPUTING SYSTEMS

at the

RHEINLAND-PFÄLZISCHE TECHNISCHE UNIVERSITÄT

January 2025

© 2025 Sean Gloumeau.

| | |
|---|---|
| Authored by: | Sean Gloumeau |
| | European Masters in Embedded Computing Systems |
| | January 22, 2025 |
| Einprüfer: | Prof. Dr.-Ing. Wolfgang Kunz |
| | Professor of Electrical Engineering and Computer Science |
| Zweitprüfer: | Prof. Dr. Marius Kloft |
| | Professor of Computer Science, Thesis Supervisor |

# Eigenständigkeitserklärung

Ich versichere hiermit, dass ich die vorliegende Masterarbeit mit dem Thema „Post-hoc Calibrated Anomaly Detection" selbstständig, d.h. auch *ohne* Verwendung von textgenerierender KI und KI-Schreibtools (ChatGPT etc.) verfasst und keine anderen als die angegebenen erlaubten Hilfsmittel benutzt habe. Alle Stellen und Gedanken meiner Arbeit, die wortwörtlich oder dem Sinn nach anderen Werken einschließlich Internetquellen entnommen sind, wurden unter Angabe der Quelle kenntlich gemacht.

Kaiserslautern, 22.01.2025

_________________________________                    _________________________________
(Ort, Datum)                                                              (Verfasser)

# Post-Hoc Calibrated Anomaly Detection
## *Post-hoc kalibrierte Anomalieerkennung*

by

Sean Gloumeau

Submitted to the Fachbereich Informatik
on January 22, 2025 in partial fulfillment of the requirements for the degree of

EUROPEAN MASTERS IN EMBEDDED COMPUTING SYSTEMS

## ABSTRACT

Deep unsupervised anomaly detection has seen improvements in a supervised binary classification paradigm in which auxiliary external data is included in the training set as anomalous data in a process referred to as outlier exposure, which opens the possibility of exploring the efficacy of post-hoc calibration for anomaly detection and localization. Post-hoc Platt scaling and Beta calibration are found to improve results with gradient-based input perturbation, as well as post-hoc training with a strictly proper loss of a base model initially trained on an unsupervised loss. Post-hoc calibration is also found at times to be more effective using random synthesized spectral data as labeled anomalous data in the calibration set, suggesting that outlier exposure is superior only for initial training.

Thesis supervisor: Prof. Dr. Marius Kloft
Title: Professor of Computer Science

# Contents

# List of Figures

# List of Tables

# Chapter 1

# Introduction

Anomaly detection aims to correctly detect data that significantly deviates from a set of normal data, and is often framed as an unsupervised learning one-class classification task with the goal of recognizing low-probability events with acess to only the normal data. It has important applications for a broad set of industries including finance, healthcare, and cybersecurity, through detection of fraud, system failures, network intrusions, and even novel events that may lead to new scientific discoveries.

Calibration, on the other hand, is a property that is defined strictly in a supervised setting for estimating conditional probabilities. Models that are calibrated produce probability estimates that have strong empirical basis, leading to results that more accurately quantify the uncertainty of a prediction.

While these two concepts are seemingly at odds with each other, seminal theoretical results have facilitated the exploration of their intersection, which is precisely the aim of this thesis. Specifically, the focus is on calibrating an initially-trained model, referred to as *post-hoc calibration.*

## 1.1   Previous Work

Calibration in the supervised setting is typically explored in relation to uncertainty quantification, and experienced a wave of popularity in the machine learning literature from a landmark paper by Guo et al. [20] with the introduction of the post-hoc calibration method temperature scaling. Since then, calibration has played an increased role in the performance evaluation of large language models [13, 45, 55].

Temperature scaling has been explored for the supervised out-of-distribution detection task, with the goal of detecting samples at test time that don't belong to any of the classes in the training set. ODIN [31] shows that temperature scaling with gradient-based input perturbation is effective for improving performance. Adaptive temperature scaling [51] develops a technique for supplying an input-dependent temperature for neural networks during inference through consideration of the empirical distributions of each layer's mean activation over a calibration set after initial network training. The mean activations of a test input are compared against these empirical distributions to determine $p$-values per layer, which are aggregated by Fisher's method to yield the temperature.

However, there is a much broader literature on calibration, with prominent post-hoc methods including parameterized methods such as Platt scaling [39] and Beta calibration [28], and binning methods including histogram binning [52], isotonic regression [53], and Bayesian binning into quantiles [38]. The underlying theory extends as far back as 1950 in the meteorological science literature with the inception of the Brier score [6].

The work of Menon and Williamson [36] stands out for exploring the sparse intersection of calibration and unsupervised anomaly detection. They extend Steinwart's work to develop a general modification of strictly proper losses, that share a deep relationship with calibration, for training models that provide a confidence score of a detected anomaly being anomalous, a task termed by them as calibrated anomaly detection.

Conformal prediction is a related technique that improves uncertainty quantification [44, 48], which has an associated analogue in anomaly detection known as, fittingly, conformal anomaly detection [21, 24, 29]. This involves mapping data in a calibration set to scores over which ranking-based statistics can be performed, facilitating the computation of a specified quantile which defines the anomaly threshold, and a $p$-value for test inputs to test against the threshold. Given the close relationship between calibration and conformal prediction, it's unsurprising Deng et al.'s work [14], that mainly makes use of concepts from conformal prediction rather than calibration, centers calibration in its discussion.

## 1.2 Outline

The thesis begins to explore post-hoc calibrated anomaly detection in a methodical way by first defining calibration and exploring related concepts in isolation of anomaly detection in chapter 2. Chapter 3 then proceeds to outline theoretical work that facilitates application of concepts from calibration to anomaly detection, and builds on this by proposing post-hoc calibration methods for anomaly detection. The body concludes with a report and discussion on experimental methodology and results in chapter 4.

# Chapter 2

# Calibration and Strictly Proper Losses

*Calibration*, also referred to as *reliability*, is a desirable property of any *forecaster* optimized for *class probability estimation*. I begin to dissect this statement in section 2.1 by defining class probability estimation (CPE), which introduces the notion of a *loss* over which a family of forecasters is optimized which builds the terminology required for the definition of *perfect calibration* in section 2.2. Section 2.3 expounds on losses in general, leading to a canonical decomposition of losses containing a *calibration term* that motivates the introduction of *(strictly) proper losses* as the natural losses for CPE that encourage calibration.

I close this chapter by restricting to the binary case in section 2.4, introducing *binary composite losses* that facilitate extending a CPE loss to neural networks, introducing stationarity conditions for CPE and composite losses that can be used as a tool to check for propriety, and defining two popular strictly proper losses, the log and logistic losses.

## 2.1 Class Probability Estimation (CPE)

**DEFINITION 1: CLASS PROBABILITY ESTIMATION.** Given data space $\mathcal{D} := \mathcal{X} \times \mathcal{Y}$ of an independent input $X \in \mathcal{X}$ and dependent output $Y \in \mathcal{Y} \subseteq \mathbb{N}$, distributed according to joint distribution $\mathbb{P}$ from which $n$ samples are drawn to form dataset $D_n := \{(X, Y)\}_n \sim \mathbb{P}^n$, class probability estimation seeks to provide estimates $\hat{\gamma}$ of the conditional distribution $\gamma := \mathbb{P}_{Y|X}(y|x)$ for all observable inputs.

Before moving on, there are a few points that should be made. First by definition, the conditional distribution and its estimate are functions $\gamma, \hat{\gamma} : \mathcal{Y} \to [0, 1]$ which map each $y \in \mathcal{Y}$ to probability value $\hat{\gamma}(y) \in [0, 1]$. Second, note that due to Bayes' rule, after introducing $M := \mathbb{P}_X(x)$, the marginal distribution of $X$, one can completely specify an experiment using either $\mathbb{P}$ or $(\gamma, M)$.

### 2.1.1 The Empirical Approach

A classical approach to estimating $\gamma$ for a given input $x$ involves directly quoting the empirical conditional probability distribution over $D_n$. Denoting the set of data pairs with the same input $x$ as $D_x = \{(x_i, y_i) : \ x_i = x, \ (x_i, y_i) \in D_n\}$, and $\mathbf{I}[p]$ as the indicator function for predicate $p$, this can be expressed as

$$\hat{\gamma}(y|x) = \frac{1}{|D_x|} \sum_{y_i \in D_x} \mathbf{I}\big[y_i = y\big]$$

While $\hat{\gamma}$ becomes a better estimate as $n \to \infty$, it becomes exponentially worse as the dimensions of the dataspace increase due to the curse of dimensionality. In fact, the curse of dimensionality poses a much bigger problem: as $\mathcal{D}$ increases in size or dimension, it becomes much less likely that $D_n$ contains all observable inputs. To see why this is a problem, note that for any input such that $x \notin D_n$, $\hat{\gamma}$ is undefined as $D_x = \varnothing$.

We could solve this problem through Parzen windowing to extrapolate a distribution over all of $\mathcal{D}$ from $D_n$, but towards the more modern estimation methods originating from statistical learning theory, one can propose an arbitrary function $g : \mathcal{X} \to \mathcal{V}$ on the input space and consider a $\gamma$-estimate conditioned on the value of $g(x) = v \in \mathcal{V}$ rather than $X$. For instance, $g$ could represent the action of binning the input space such that each bin contains at least one $x \in \mathcal{X}$, or it could also represent *feature extraction*. The latter has proven useful in computer vision for object detection, where features such as edges, vertices or color masks can be extracted via popular computer vision transformations.

Similarly to above, let the set of data pairs with the same transformed input $g(x) = v$ be denoted as $D_v = \{(x_i, y_i) : \ g(x_i) = v, \ (x_i, y_i) \in D_n\}$; the estimate $\hat{\gamma}$ is given as

$$\hat{\gamma}(y|v = g(x)) = \frac{1}{|D_v|} \sum_{y_i \in D_v} \mathbf{I}\big[y_i = y\big] \tag{2.1}$$

With the current estimation approach it's evident that $g$ only prevents undefined estimates when it bins in some way. What if instead of quoting an empirical estimate over $D_n$, we instead find some function defined over all of $\mathcal{X}$ by *training on $D_n$* that provides an estimate for any $x \in \mathcal{X}$?

### 2.1.2 Forecasting

Let's now turn our attention to another approach for estimating $\gamma$. The goal is to find a *forecaster* $\hat{f}_\theta : \mathcal{X} \to \mathcal{P}_\mathcal{Y}$ among a family $F_\theta$ of $\theta$-specified forecasters that map an input $x$ to a conditional distribution estimate, $f_\theta \in F_\theta : f_\theta(x) \mapsto \hat{\gamma}$, where $\mathcal{P}_\mathcal{Y}$ denotes the family of all distributions over $\mathcal{Y}$. This is achieved by optimizing against a *loss* $\ell(y, \hat{\gamma})$ in the sense of equation 2.4, a process also referred to as training a forecaster on $D_n$, and since $D_n$ is a random variable it is evident that so too is the forecaster. To drive this point home: it is not unreasonable to expect data drift over the course of years, and so naturally the optimal forecaster found by training over two separate datasets randomly sampled at different points in time should differ to adapt to this shift, hence it is desirable for $\hat{f}_\theta$ to be modeled as a random variable dependent on $D_n$.

## 2.2 Calibration

Consider random forecasters $f_\theta$ trained over dataset $D_n \sim \mathbb{P}^n$ to predict $\hat{\gamma}$ as an estimate of $\gamma$. A desirable property of the forecaster's estimates is that they match the empirical estimates of equation 2.1 in expectation over $\gamma$. *Calibration* requires this matching when conditioned on a choice of $v = f_\theta(x)$[1].

**DEFINITION 2: PERFECT CALIBRATION [9].** A forecaster $f_\theta : \mathcal{X} \to \mathcal{P}_\mathcal{Y}$ trained to estimate conditional mass $\gamma$ is said to be perfectly calibrated if for all $\hat{\gamma} : \mathcal{Y} \to [0, 1]$ in the range of $f_\theta$ and $y \in \mathcal{Y}$

$$\Pr_{Y \sim \gamma} \big( Y = y | f_\theta(X) = \hat{\gamma} \big) = \hat{\gamma}(y). \tag{2.2}$$

The term on the left is interpreted in a frequentist manner

$$\Pr_{Y \sim \gamma} \big( Y = y | f_\theta(X) = \hat{\gamma} \big) = \mathbb{E}_{Y \sim \gamma} \big[ \mathbf{I}[Y = y] | f_\theta(X) = \hat{\gamma} \big]$$

such that if a perfectly calibrated forecaster predicting a probability $\hat{\gamma}(y)$ for some outcome $y$ given an input $x$, then that outcome occurs $\hat{\gamma}(y)$ percent of the time as time tends to infinity [11].

The forecaster output $\hat{\gamma}$ is termed a *subjective probability* [12], which emphasizes the notion that inference is performed even in situations where objective data is not present; i.e., inference generalizes to unseen inputs. Calibration can thus be seen as a desirable property for this subjectivity to reflect the objective empirical observations when present in the data, intuitively as some foundational pillar from which the forecaster can then generalize to provide its subjective opinion.

## 2.3 Losses

**DEFINITION 3: LOSSES.** For a set of output values $\mathcal{Y}$, prediction space $\mathcal{V}$ and codomain $B \subset \mathbb{R}$ for which an infimum exists, a loss is any function $\ell : \mathcal{Y} \times \mathcal{V} \to B$ designed to incur a penalty for the deviation of a prediction $v$ from an observed output $y$, with a larger penalty incurred for a larger deviation.

Note that translating $B$ by (the additive inverse of) its infimum induces a more general codomain with infimum 0, $R_{\geq 0} := [0, \infty) \subset \mathbb{R}$, that contains $B$, hence w.l.o.g. the codomain of $\ell$ is often denoted as $R_{\geq 0}$. Losses defined in this way frame optimization as loss minimization problems, while losses defined dually over a codomain for which a supremum exists lead to loss maximization problems. *Unsupervised losses* are recovered from this definition

---

[1]There are other similar notions of calibration in the broader literature, in particular *probabilistic* and *marginal* calibration, however this thesis focuses on what is more precisely known as *conditional calibration* which is prevalent and referred to as simply calibration in the machine learning literature. The curious reader is referred to [16, 18] for more information on other notions

for the special case of singleton output sets $\mathcal{Y} = \{*\}$, for which $\ell$ does not effectively depend on the value of $y$, *binary losses* are recovered for $|\mathcal{Y}| = 2$, *multiclass losses* for $\mathcal{Y} \subseteq \mathbb{N}$ and $|\mathcal{Y}| > 2$, and *regression losses* if none of the prior criteria apply and $\mathcal{Y} \subseteq \mathbb{R}^n$.

In the context of optimization, predictions $v$ are the output of any *predictor* $g : \mathcal{X} \to \mathcal{V}$ acting on input $x$ that's dependent on a dataset $D_n \sim \mathbb{P}^n$, and is hence modeled as a random variable similarly to forecasters. In the case of CPE where $\mathcal{V} = \mathcal{P}_{\mathcal{Y}}$ and $g = f_\theta \in F_\theta$ is a forecaster, one can intuit that well-designed losses penalize conditional distributions that are not concentrated about the observed output.

### 2.3.1 Risks: Losses in Expectation

Predictors are not trained to minimize directly against losses, but rather against *the expectational loss with respect to* $\mathbb{P}$.

**DEFINITION 4: FULL RISK.** Let $(X, Y)$ be random variables distributed according to joint distribution $\mathbb{P}$, with associated conditional and marginal densities $\gamma$ and $M$ respectively. The full risk of a predictor $g : \mathcal{X} \to \mathcal{V}$ with respect to a loss $\ell(y, v)$ is defined as

$$\mathbb{L}(\mathbb{P}, v = g(X)) := \underset{(X,Y)\sim\mathbb{P}}{\mathbb{E}}[\ell(Y, g(X))] = \underset{X\sim M}{\mathbb{E}}\Big[\underset{Y\sim\gamma}{\mathbb{E}}[\ell(Y, g(X))|X = x]\Big] \qquad (2.3)$$

A family of quasi-differentiable predictors parameterized by $\theta \subseteq \mathbb{R}^n$ is typically proposed over which the full risk is evaluated for the purposes of minimization. These predictors are amenable to automatic differentiation, and as a result have the benefit of being specified as the action of a neural network with weights $\theta$ that can be optimized by updating weights via gradient descent to minimize the full risk. With this example in mind, more generally for some family of predictors $G_\theta$ specified by $\theta$ that's a subset of all measurable functions $G$ from $\mathcal{X}$ to $\mathcal{V}$, the *optimal predictor* $\hat{g}_\theta \in G_\theta$ is

$$\hat{g}_\theta := \underset{g_\theta \in G_\theta}{\arg\inf} \mathbb{L}(\mathbb{P}, g_\theta(x)) \qquad (2.4)$$

The smallest possible full risk *over all of $G$* is referred to as the *Bayes risk* $\mathbb{L}^*(\mathbb{P})$,

$$\mathbb{L}^*(\mathbb{P}) := \underset{g \in G}{\inf} \mathbb{L}(\mathbb{P}, g(x)) \qquad (2.5)$$

and the predictor for which this is achieved the *Bayes optimal predictor* $g^*$,

$$g^* := \underset{g \in G}{\arg\inf} \mathbb{L}(\mathbb{P}, g(x)) \qquad (2.6)$$

Another risk of import is based on the *conditional expectational loss*.

**DEFINITION 5: CONDITIONAL RISK.** Given prediction space $\mathcal{V}$ and a random variable $Y$ distributed according to conditional distribution $\gamma$, the conditional risk of loss $\ell(y, v)$ for prediction $v \in \mathcal{V}$ is defined as

$$L(\gamma, v) := \underset{Y\sim\gamma}{\mathbb{E}}[\ell(Y, v)] \qquad (2.7)$$

In the same spirit of the Bayes risk, the smallest possible conditional risk over all of $\mathcal{V}$ is referred to as the *conditional Bayes risk* $L^*(\gamma)$:

$$L^*(\gamma) := \inf_{v \in V} L(\gamma, v) \tag{2.8}$$

Note that conditional risks are special in the sense that with only $\mathcal{V}$, one can reason about it and its underlying loss without consideration of inputs or predictors, making it a useful tool for the design of losses. Nevertheless, it is still perfectly sound to reason about conditional risk in the context of predictors as random variables $g : x \mapsto v$ dependent on $D_n$, in which case the conditional risk becomes a component of the full risk in equation 2.3: $L(\gamma, v = g(x)) = \underset{Y \sim \gamma}{\mathbb{E}} [\ell(Y, v = g(X))|X = x]$, implying that any predictor achieving the conditional Bayes risk for all $x \in \mathcal{X}$ achieves the Bayes risk.

## 2.3.2 (Strictly) Proper Losses

Here's a simple decomposition of the conditional risk where, in true statistical learning theory fashion, a term is simply added and subtracted [7, 8]:

$$L(\gamma, v) = \underbrace{L(\gamma, \gamma)}_{L_e,\text{ entropy term}} + \underbrace{L(\gamma, v) - L(\gamma, \gamma)}_{L_c,\text{ calibration term}} \tag{2.9}$$

The *entropy term* $L_e$, also referred to as the *generalized entropy function of* or *information measure of* $\ell$ for $\gamma$, only depends on the conditional distribution, while the *calibration term* $L_c$ also depends on the prediction space $\mathcal{V}$. Since the latter accommodates predictions, it becomes the focus for defining properties of loss functions.

**DEFINITION 6: (STRICTLY) PROPER LOSSES [17].** A loss $\ell(y, v)$ is said to be *proper* if $L_c(\gamma, v) \geq 0 \quad \forall v \in \mathcal{V}$ and *strictly proper* if, in addition, $L_c(\gamma, v) = 0$ iff $v = \gamma$.

An immediate consequence of definition 6 is that all proper losses have conditional Bayes risk $L^*(\gamma) = L(\gamma, \gamma)$ since

$$L_c(\gamma, v) = L(\gamma, v) - L(\gamma, \gamma) \geq 0 \quad \forall v \in \mathcal{V}$$
$$\longrightarrow L(\gamma, v) \geq L(\gamma, \gamma) \qquad \forall v \in \mathcal{V}$$

That is, when a loss $\ell(y, v)$ is proper, by definition its Bayes conditional risk $L(\gamma, v)$ is achieved for perfect prediction of the conditional distribution $v = \gamma$, a property known as Fisher consistency, that makes these losses a natural choice for CPE with forecasters [10]. It is sensible to design forecasters trained on strictly proper losses to contain $\gamma$ in their range, which leads to the design choice of $\mathcal{V} = \mathcal{P}_\mathcal{Y}$ as the smallest set guaranteed to contain the Bayes optimal predictor.

Furthermore, any Bayes optimal forecaster that makes this perfect prediction of the conditional distribution is perfectly calibrated. One need only inspect equation 2.2 to understand that $\hat{\gamma} = \gamma$ causes both sides to equal $\gamma$. This motivates the nomenclature of calibration term for $L_c$, which is often interpreted as a deviation from perfect calibration. To sum up, when trained on a strictly proper loss, forecaster parameters are optimized to reduce $L_c$ by minimizing the conditional risk which yields better-calibrated forecasters.

**On (Strict) Propriety of Pixel-Wise Losses**

The following proposition shows a straightforward way to extend any strictly proper reference loss to a strictly proper segmentation loss by taking the average of the reference loss over pixels of the inference mask.

**PROPOSITION 1.** Consider pixel-wise classification, where $X$ is an image with $n$ pixels and $Y = (Y_i)_1^n \sim \gamma$ is a segmentation mask assigning pixel $i$ a class $Y_i$. Let $\mathbb{P}$ be a joint distribution over $\mathcal{D} = \mathcal{X} \times \mathcal{Y}$ with conditional distribution $\gamma := \mathbb{P}_{Y|X}(y|x)$, and assume $Y_i$ is conditionally independent from all $Y_{j \neq i}$ given $X$ such that pixel-wise conditional distributions are given by $\gamma_i = \mathbb{P}_{Y_i|X}(y_i|x)$. Then, the average of pixel-wise losses

$$\ell(y, \hat{\gamma}) = \frac{1}{n} \sum_{i=1}^n \ell_i(y_i, \hat{\gamma}_i)$$

is (strictly) proper if all pixel-wise losses $\ell_i(y_i, \hat{\gamma}_i)$ are (strictly) proper.

*Proof.* A proof of a similar statement for $n = 2$ and the sum rather than average can be found under proposition 2 of Appendix A. Inductive application thus shows that $\sum_{i=1}^n \ell_i(y_i, \hat{\gamma}_i)$ is (strictly) proper, and since multiplication by a constant doesn't change the minimizer of a function, the average is hence also (strictly) proper. ∎

## 2.4 The Binary Setting

The binary setting is natural for unsupervised anomaly detection, as shown in detail in section 3.1, however informally the setting is appropriate considering there are only two classes for test data: normal and anomalous. By convention, $\mathcal{Y} = \{0, 1\}$ with $y = 0$ denoting normal and $y = 1$ denoting anomalous data.

Now the conditional distribution is Bernoulli distributed $\gamma = \text{Be}(\eta)$ and is fully characterized by a single probability $\eta = \mathbb{P}_{Y|X}(y = 1|x)$, and hence the shorthand notation $Y \sim \eta$ will be used to denote $Y \sim \text{Be}(\eta)$. In addition, CPE aims to find an estimate $\hat{\eta} \in [0, 1]$ for $\eta$.

### 2.4.1 Binary CPE Losses

Binary CPE losses $\ell(y, \hat{\eta})$ can be decomposed into partial losses $\ell_0(\hat{\eta})$ and $\ell_1(\hat{\eta})$: $\ell(y, \hat{\eta}) = y\ell_1(\hat{\eta}) + (1 - y)\ell_0(\hat{\eta})$. The conditional risk then becomes

$$
\begin{aligned}
L(\eta, \hat{\eta}) &= \underset{Y \sim \eta}{\mathbb{E}} \left[ Y\ell_1(\hat{\eta}) | Y = 1 \right] + \underset{Y \sim \eta}{\mathbb{E}} \left[ (1 - Y)\ell_0(\hat{\eta}) | Y = 1 \right] \\
&= \eta\ell_1(\hat{\eta}) + (1 - \eta)\ell_0(\hat{\eta})
\end{aligned}
$$

Using the fact that the conditional risk is minimized for $\hat{\eta} = \eta$ leads to a stationarity condition for proper losses.

**DEFINITION 7: STATIONARITY CONDITION [10].** The stationarity condition ensuring a binary loss is proper is given by

$$\frac{\partial}{\partial \hat{\eta}}\bigg|_{\hat{\eta}=\eta} L(\eta, \hat{\eta}) = 0$$
$$\Rightarrow (1-\eta)\ell_0'(\eta) = -\eta\ell_1'(\eta) \tag{2.10}$$

**Log Loss**

Log loss is given by $\ell_0(\hat{\eta}) = -\ln(1-\hat{\eta})$ and $\ell_1(\hat{\eta}) = -\ln(\hat{\eta})$:

$$\ell(y, \hat{\eta}) = -y\ln(\hat{\eta}) - (1-y)\ln(\hat{\eta}) \tag{2.11}$$

A proof that log loss is strictly proper is given in Appendix A under proposition 4.

## 2.4.2 Binary Composite Losses

Binary CPE losses $\ell(y, \hat{\eta}) : \mathcal{Y} \times [0, 1] \to \mathbb{R}_{\geq 0}$ can accommodate arbitrary prediction spaces $\mathcal{V}$ through composition with the inverse of an invertible link function $\psi : [0, 1] \to \mathcal{V}$ to form what's referred to as a *composite binary losses* [40] $\lambda(y, v) = \ell(y, \psi^{-1}(\hat{\eta}))$. The core idea is that the inverse link maps predictions into a probability estimate which can be directly used by a binary CPE loss.

Similar to binary CPE losses, binary composite losses can also be decomposed into partial losses and have a stationarity condition that follows from substitution of $\ell_i'(\hat{\eta}) = (\lambda_i \circ \psi(\hat{\eta}))' = \psi'(\hat{\eta})\lambda_i'(\psi(\hat{\eta}))$ into equation 2.10:

**DEFINITION 8: STATIONARITY CONDITION FOR BINARY COMPOSITE LOSSES.** Binary composite losses are proper if they meet the stationarity condition

$$(1-\eta)\lambda_0'(\eta) = -\eta\lambda_1'(\eta) \tag{2.12}$$

Proper binary composite losses can be used to train neural networks $\phi_\theta(x) : \mathcal{X} \to \mathcal{V}$, that typically don't have outputs in $[0, 1]$, to encourage calibration. Note that the composition of the network with inverse link induces a forecaster.

**Logistic Loss**

The logistic loss used in classical logistic regression is given for $\ell(y, \hat{\eta})$ as the log loss defined in equation 2.11, and logit link function:

$$z(\hat{\eta}) = \ln(\hat{\eta}) - \ln(1-\hat{\eta}) \tag{2.13}$$

The inverse link is the sigmoid function:

$$\sigma(v) = \frac{1}{1 + e^{-v}} \tag{2.14}$$

Hence, logistic loss is given as

$$\lambda(y, v) = \ell(y, \sigma(v)) = y \ln(1 + e^{-v}) + (1 - y) \ln(1 + e^{v}) \tag{2.15}$$

A proof that logistic loss is strictly proper is given in Appendix A under proposition 3.

### 2.4.3 Measuring Calibration

First, note that the definition of perfect calibration in equation 2.2 simplifies for binary losses:

$$\Pr_{Y \sim \eta}(Y = 1 | f_\theta(X) = \hat{\eta}) = \hat{\eta} \tag{2.16}$$

Both sides of the equation can be empirically estimated through the introduction of a binning scheme. Let $K \in \mathbb{N}$ equal-width bins $B_k$ partition $[0, 1]$, with corresponding bin boundaries $(\eta_k, \eta_{k+1})$. Then when an estimate $\hat{\eta}_i$ for input $x_i$ and associated class label $y_i$ is predicted, it is placed into the $k$'th bin such that $\eta_k < \hat{\eta}_i \leq \eta_{k+1}$.

The estimate for bin $B_k$ for the left hand side of equation 2.16 is the empirical frequency of class $y = 1$:

$$\text{freq}(B_k) = \frac{1}{|B_k|} \sum_{i \in B_k} I[y_i = 1]$$

The estimate for bin $B_k$ for the right hand side of equation 2.16 is referred to as average confidence:

$$\text{conf}(B_k) = \frac{1}{|B_k|} \sum_{i \in B_k} \hat{\eta}_i$$

Calibration errors can then be derived from the absolute difference of the two [38].

**Definition 9: Max Calibration Error (MCE).** Max calibration error is given by

$$MCE = \max_k(|\text{freq}(B_k) - \text{conf}(B_k)|)$$

**Definition 10: Expected Calibration Error (ECE).** Expected calibration error is given by

$$ECE = \sum_{k=1}^{K} \frac{B_k}{n} (|\text{freq}(B_k) - \text{conf}(B_k)|)$$

# Chapter 3

# Post-Hoc Calibration for Anomaly Detection

With the basics of calibration and proper losses out of the way, this chapter swiftly proceeds to discuss calibration for anomaly detection. In section 3.1, an overview of theoretical results that justifies calibration for anomaly detection is provided, along with consideration of positive outlier exposure results from the perspective of calibration.

Section 3.2 is the main highlight of this chapter, and presents three post-hoc calibration methods that will be evaluated in chapter 4. Motivated by the lack of two of these methods having the capability to permute anomaly score rankings, section 3.3 introduces gradient-based input perturbation of test samples. The chapter concludes with a discussion of synthetic anomalous data in section 3.4, and outlines a form of random image synthesis used in post-hoc calibration experiments.

## 3.1 Calibrated Anomaly Detection

Unsupervised anomaly detection can be framed as the *density level detection* problem, which aims to estimate the level sets $\{x \in \mathcal{X} : m(x) > \rho, \rho > 0\}$ of the marginal density $m$ (the derivative of marginal distribution $M$ as defined in section 2.1) for a given density threshold $\rho$, with the core idea that anomalies are classified as rare, low-density events that don't belong to the level set for a given threshold.

Seminal work by Steinwart et al. [46] showed that binary classifiers trained to classify normal data from random synthetic anomalies drawn from some known reference density are asymptotically consistent density level set estimators for an appropriately cost-weighted $0 - 1$ loss ([46], Proposition 5), in the sense that the target for DLD can be recovered from classification. Menon and Williamson [36] extend this result to so-called classification-calibrated losses [3], and show that the target for classification can be recovered from $\hat{\eta}$, the target of CPE, for strictly proper binary composite losses using the associated link function. That is, strictly proper losses can be used as surrogates for DLD, which is referred to as *calibrated anomaly detection* [36] [1].

---

[1]Note that Menon and Williamson restrict the term calibrated anomaly detection to the task of partial density estimation, however I appropriate the term here for any application of proper scoring rules to anomaly

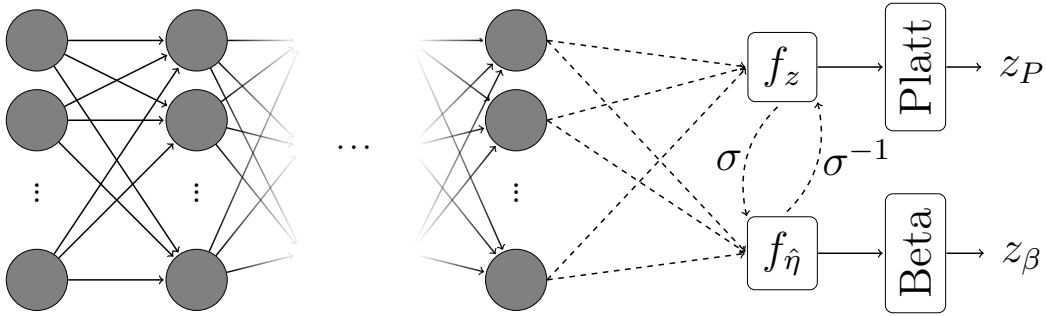

Figure 3.1: Platt scaling and $\beta$ calibration. The base network is frozen, with the output being used to directly compute either logits $z$ or estimates $\hat{\eta}$ with $f_z$ or $f_{\hat{\eta}}$ respectively, from which the other can be determined using either the sigmoid $\sigma$ or logit $\sigma^{-1}$ functions respectively. These values are then used as input to optimize Platt or $\beta$ parameters with the logistic function.

This is precisely what opens the opportunity to explore calibration in the unsupervised setting for anomaly detection; one can induce a classification setting, in which a strictly proper loss is used as surrogate, by labeling all available normal training data with $y = 0$, and drawing random data from another source to produce synthetic anomalies labeled as $y = 1$.

Strictly proper losses can also be used as surrogates for the bipartite ranking problem, whose full risk is given by $1 - \text{AUROC}$, justified through a surrogate risk bound [1] that shows optimization of strictly proper losses leads to improved performance of bipartite ranking. Hence we can expect that training on a strictly proper loss leads to rankings of estimate $\hat{\eta}$ that improve the AUROC.

Let us now turn our attention to experiments in [35] related to *outlier exposure*, whereby synthetic anomalous data for images is sourced from auxiliary, external datasets to induce classification as a surrogate task for anomaly detection. It was shown that logistic loss, which is strictly proper composite, and hypersphere classifier loss (cf. section 4.1.2), which uses the strictly proper log loss, outperform other classical unsupervised anomaly detection losses; i.e. calibrated anomaly detection produced the best results. It can thus be reasonably hypothesized that calibration is a desirable property for anomaly detection.

## 3.2  Post-Hoc Calibrated Anomaly Detection

Post-hoc calibration is a technique making that splits the training data to accommodate two separate training phases in order to improve calibration of a model. The initial training phase trains the full model, while the second one, referred to as the calibration phase, freezes the model to drastically reduce the parameter space, and optimizes over a strictly proper scoring rule. For a model initially trained on an unsupervised loss for anomaly detection, it is plausible that the calibration phase should improve results based on the discussion on outlier exposure in the previous section, however it is not as clear why it would improve detection.

results for models initially trained on a strictly proper loss with outlier exposure.

To gain more insight, we turn to the machine learning literature on calibration for supervised classification with deep neural networks. The negative log likelihood, which is strictly proper composite, is typically used as optimization loss and should lead to calibrated estimates, however neural networks tend to be overconfident in their predictions. This primarily happens for the following reasons [20]:

- Models have increasingly large model capacity that leads to overfitting on the negative log likelihood,

- Reduced weight decay and regularization for optimizers in modern networks,

- Batch normalization, which has been shown to empirically decrease calibration.

All three causes lead to improvements in accuracy, but at the cost of producing well-calibrated estimates. Hence, post-hoc calibration can be helpful in high-risk applications where overconfident predictions can't be afforded.

For anomaly detection, it may be possible that calibrating a network initially trained on a strictly proper loss could see improved AUROC scores simply due to ranking permutations of the anomaly score as the model becomes more measured in its predictions. Three different post-hoc calibration methods are explored in this work:

1. Platt scaling,

2. Beta / $\beta$ calibration,

3. Modifying the network to end with a fully connected layer to one perceptron, and freezing weights of all but the final layer, which I refer to as the *calibration head* method.

The first two methods, illustrated in figure 3.1, are well-established in the literature. The last was designed to facilitate ranking permutations without requiring anything extra (cf. input perturbation in section 3.3).

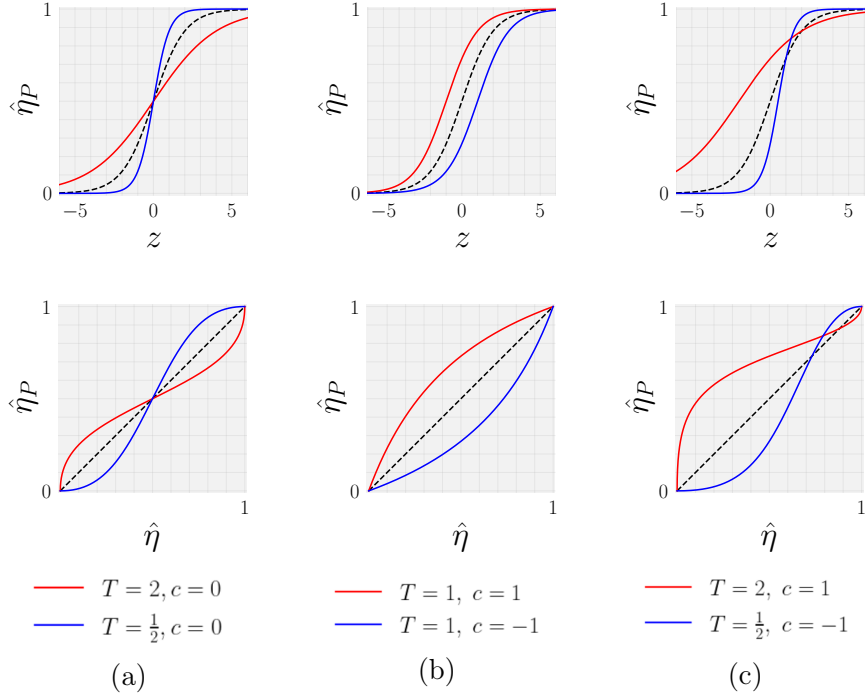

Figure 3.2: The effect of Platt scaling on calibrated probability estimates $\hat{\eta}_P$ with respect to the original logits $z$ and estimate $\hat{\eta}$. Identity mappings for $T = 1$, $c = 0$ are shown with a dashed black line. (a) $T$ affects symmetric sigmoid dilation about the intercept; (b) $c$ translates the sigmoid intercept; (c) Parameter combinations lead to a rich set of estimate transformations.

### 3.2.1 Platt Scaling

Platt scaling [39] involves two parameters, temperature $T \in \mathbb{R}_{\geq 0}$ and intercept $c \in \mathbb{R}$, used to create a simple single-variable linear transformation of logits $z$ derived from neural network output:

$$z_P = \frac{z}{T} + c \tag{3.1}$$

The calibrated probability estimate $\hat{\eta}_P$ is computed as the sigmoid of the transformed logit:

$$\hat{\eta}_P = \sigma(z_P) = 1 \left/ (1 + \exp(-\frac{z}{T} - c)) \right.$$

The parameters are optimized through minimization of logistic loss using $z_P$ as inputs. $T$ is primarily responsible for *sigmoid dilation* and $c$ for *intercept translation*, as shown in figure 3.2.

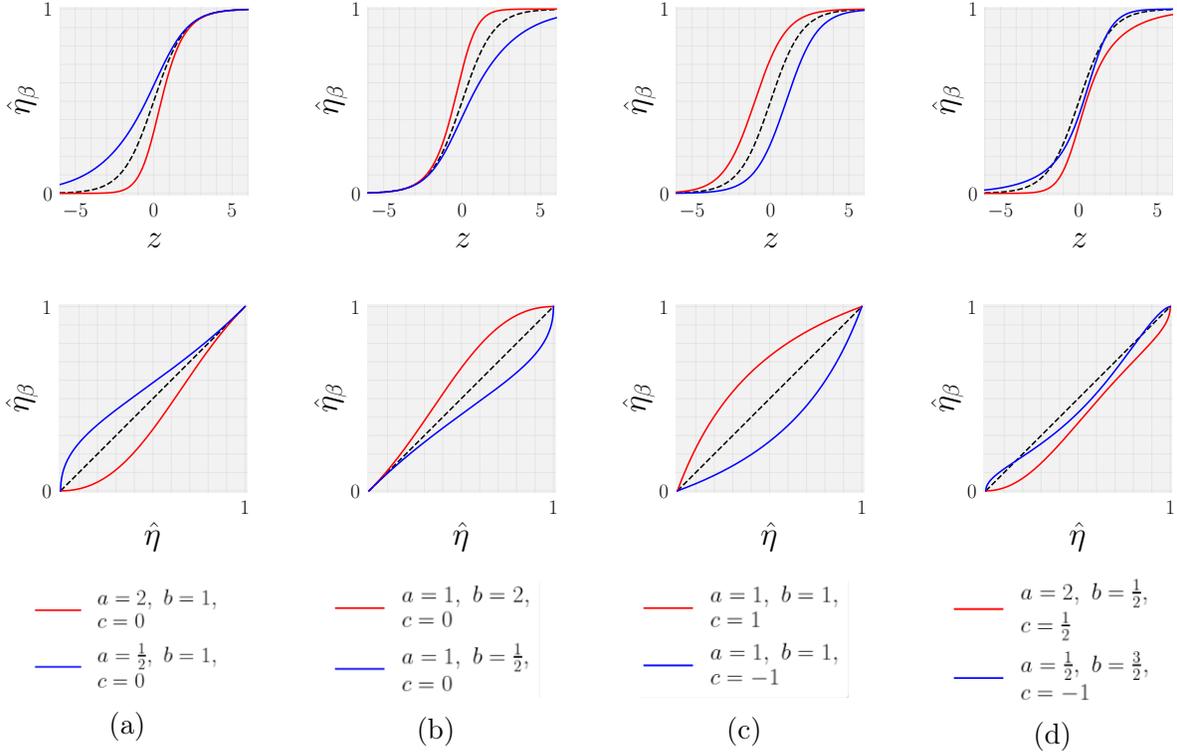

Figure 3.3: The effect of Beta calibration on calibrated probability estimates $\hat{\eta}_\beta$ with respect to the original logits $z$ and estimate $\hat{\eta}$. Identity mappings for $a = b = 1$, $c = 0$ are shown with a dashed black line. (a) $a$ affects left-tailed sigmoid dilation; (b) $b$ affects right-tailed sigmoid dilation; (c) $c$ translates the sigmoid intercept; (d) Parameter combinations lead to a rich set of estimate transformations.

### 3.2.2 Beta Calibration

Beta / $\beta$ calibration [28] involves 3 parameters $a$, $b \in \mathbb{R}_{\geq 0}$ and $c \in \mathbb{R}$ used to transform the probability estimate $\hat{\eta}$ derived from neural network output into logits $z_\beta$:

$$z_\beta = a\ln(\hat{\eta}) - b\ln(1 - \hat{\eta}) + c \tag{3.2}$$

Note that this transformation is simply a weighted and translated form of the logit function $\sigma^{-1}(\hat{\eta})$ (cf. equation 2.13). Similarly to Platt scaling, the calibrated probability estimate $\hat{\eta}_\beta$ is computed as the sigmoid of the transformed logit:

$$\hat{\eta}_\beta = \sigma(z_\beta) = 1 \left/ \left( 1 + \exp(-c) \frac{(1 - \hat{\eta})^b}{\hat{\eta}^a} \right) \right.$$

The parameters are, again, optimized through minimization of logistic loss using the transformed logits $z_\beta$ as inputs. $a$ and $b$ are mainly responsible for sigmoid dilation of the left and right tail respectively, and $c$ is mainly responsible for intercept translation, as shown in figure 3.3.

For $a = b$, equation 3.2 becomes $z_\beta = a(\sigma^{-1}(\hat{\eta})) + c = az + c$, which is identical to the logit transformation of Platt scaling in equation 3.1 for $a = 1/T$. In other words, the $\beta$ calibration family of transformations is a superset of the Platt scaling family of transformations; in this sense, $\beta$ calibration is a generalization of Platt scaling. Intuitively, setting $a = b$ amounts to having the same dilation magnitude at both tails of the sigmoid function, causing symmetric dilation about some intercept as $T$ does for Platt scaling.

### 3.2.3    Calibration Head

The calibration head method modifies networks in the following way after training:

- For autoencoders that output to a matrix space $\phi_\theta : \mathcal{X} \to \mathbb{R}^{m \times n}$, the network is decapitated from the bottleneck layer, from which it is then fully connected to a single node.

- For networks that output to a vector space $\phi_\theta : \mathcal{X} \to \mathbb{R}^n$, the head perceptrons are fully connected to a single node.

- For networks that output to a scalar space $\phi_\theta : \mathcal{X} \to \mathbb{R}$, the final fully-connected layer's weights were reset.

After modification, all layers save for the final fully-connected layer are frozen. This technique is designed to complement Platt scaling and $\beta$ calibration by providing more free parameters for which the logistic loss can be optimized. It can also be thought of as training on features extracted by the model after the training phase, in the form of the latent output preceding the final layer. Since no nonlinear activation is applied to the final node, evaluating this method after calibration explores whether these features are linearly separable.

## 3.3    Gradient-Based Input Perturbation

While Platt scaling and $\beta$ calibration lead to improved calibration, their associated transformations are strictly increasing, leading to identical rankings of anomaly scores over the test set between the calibrated model and base network. As a result, neither would lead to an improvement on ranking-based metrics central to assessing performance of anomaly detection and localization, such as the AUROC or AUPRO.

One way to change the underlying rankings for any model, calibrated or not, is gradient-based input perturbation as described in ODIN [31], where inference is performed over a perturbed test input $\tilde{x}_t$. Let $x_t$ be the original test input, $\epsilon \in \mathbb{R}_{>0}$ be some positive constant, $\ell(y, x)$ be the loss expressed as a function of input $x$, which involves the action of a neural network and any potential postcomposing functions, and sgn denote the sign function. The perturbed input is given as

$$\tilde{x}_t = x_t - \epsilon \, \text{sgn}(\nabla_x \ell(y, x)) \tag{3.3}$$

This method is directly inspired by the fast gradient sign (FGS) method for data augmentation [19], which differs from equation 3.3 only by the replacement of the subtraction operation with an addition. The reason is that FGS seeks new training inputs that *increase* the loss, providing the network a greater challenge for improved learning and generalization on test inputs. On the other hand, input perturbation for improving anomaly detection modifies test samples in such a way to *decrease* the loss.

It has been empirically shown that small, imperceptible perturbations on inputs can have a large impact on resultant inference scores in a supervised setting with minimal to no anomalous data in the training or test sets [37, 47]; on the other hand for out-of-distribution data, ODIN empirically shows that perturbation has less impact. In other words, while input perturbation as described by equation 3.3 can cause both anomalous and normal data to look more normal to a model by decreasing the loss, in such cases *perturbed normal data looks much more normal than perturbed anomalous data*, resulting in improved separability. This is the main mechanism by which anomaly score rankings can be improved by input perturbation.

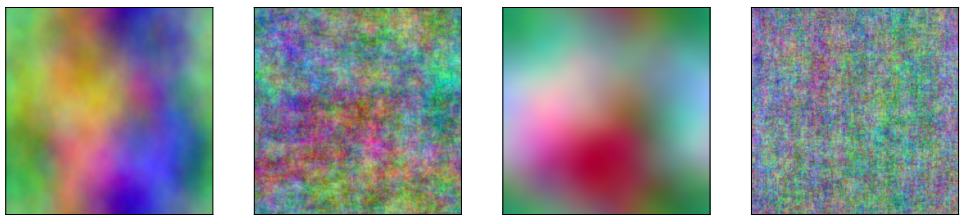

Figure 3.4: Representative samples of random data synthesized from the $1/f^\alpha$ spectra of natural images.

## 3.4 On Synthetic Anomalous Data

Traditionally, synthetic anomalous data is sampled from a uniform distribution [46]. However, it has been empirically shown that for images, training with synthetic anomalies produced from the uniform distribution, in addition to a variety of other random synthesis strategies, lead to impoverished results [33]. The leading hypothesis, in light of the success of choosing synthetic anomalies from an outlier exposure dataset, is that the dimension of the input space is too large for randomly synthesized data to properly fill it, and is thus not as informative to the network for finding a separation boundary as outlier exposure data which, while less of the input space is explored, is more structurally similar to unseen anomalous data while still maintaining sufficient diversity from the normal data.

With post-hoc calibration, however, the majority of model parameters are kept frozen in the calibration phase, meaning inputs to the parameterized functions or network layers are in a drastically reduced input space. In this way, post-hoc calibration recovers the traditional lower-dimensional setting in the calibration phase for which uniform noise has proven effective, and hence it is plausible that random data would work as well as outlier exposure. It may even perhaps be more effective than outlier exposure data, as it provides larger diversity of the reduced input space.

This hypothesis is tested with spectrally synthesized data that follow the $1/f^\alpha$ spectra of natural images [23]. More specifically, the same synthesis scheme of [2] is employed, whereby the frequency magnitude follows $1/(|f_x|^a + |f_y|^b)$ with $a, b$ uniformly sampled from $[0.5, 3.5]$ and the phase is derived from the phase of an image with each pixel uniformly distributed in RGB space $[0, 255]$. The inverse Fourier transform was then applied to yield a random spectral image. Representative samples can be inspected in 3.4

I note here that in the early experimental phase of this thesis, initial calibrated training with spectrally synthesized data was tested, and led to the same impoverished results observed in the past. The network from this training showed little discriminative power, to the point that post-hoc calibration was wholly ineffective.

# Chapter 4

# Experiments

Experiments were conducted to test the post-hoc calibrated methods discussed in section 3.2 for anomaly detection. Experiments for anomaly localization were also conducted, facilitated by proposition 1. The base losses for initial training are reviewed in section 4.1, and methodological details such as the networks, datasets, and optimization strategies are detailed in section 4.2. Finally, results are shown and discussed in section 4.3.

## 4.1 Base Losses

This section briefly outlines the base losses used for initial network training in experiments. The output space for networks trained on each loss is specified, to identify the calibration head strategy of section 3.2.3 for post-hoc calibration. Each loss has an associated logit $z$ and probability estimate $\hat{\eta}$ used as inputs for Platt scaling and $\beta$ calibration respectively, and an anomaly score used for evaluation of fully-trained networks. Note that logistic loss is not covered, as it was already defined in equation 2.15, however we note that this is the only loss for which the network output space is scalar: $\phi_\theta(x) \in \mathbb{R}$.

### 4.1.1 Support Vector Data Description (SVDD)

SVDD [42] is an unsupervised loss given by

$$\ell(x) = ||\phi_\theta(x) - \mathbf{c}||^2$$

where $\mathbf{c} \in \mathbb{R}^n$ is a hypersphere center and $||\cdot||$ is the 2-norm Euclidean distance.

This loss is geometrically motivated, as it encourages networks trained on it to map the normal training data as close to the hypersphere center as possible, leading to shorter and larger square distances for normal and anomalous data, respectively, during inference. The anomaly score is, naturally, this distance: $\mathrm{AS} = ||\phi_\theta(x) - \mathbf{c}||^2$.

SVDD has a degenerate solution referred to as *hypersphere collapse* whereby the network learns a constant mapping to the hypersphere center, which must be discouraged by choosing a center that isn't at the origin, and using networks with no bias units. Hence for all experiments, networks trained on SVDD have no bias, and I follow the author of the original paper by setting $\mathbf{c}$ to the mean of an initial forward pass of the training data.

The logit for this loss was chosen to be the anomaly score, and the corresponding probability estimate is computed from it using the sigmoid function.

### 4.1.2  Hypersphere Classifier (HSC)

HSC [35, 43] is a supervised loss for networks with vector output space $\phi_\theta(x) \in \mathbb{R}^n$ that can be expressed as a binary composite loss (cf. section 2.4.2) with inverse link function $\psi^{-1} : \mathbb{R}_{\geq 0} \to [0, 1]$, $\psi^{-1} : v \mapsto 1 - e^{-v}$, and log loss as CPE loss:

$$\lambda(y, v) = -y \ln(1 - e^{-v}) + (1 - y)v \tag{4.1}$$

HSC is motivated as a supervised version of SVDD. To that end, in the literature it is normally presented with $v = ||\phi_\theta(x)||^2$, since for normal data the SVDD loss is recovered for $\mathbf{c} = 0$: $\lambda(y = 0, v = ||\phi_\theta(x)||^2) = ||\phi_\theta(x)||^2$. In practice, however, $v$ is typically chosen as the pseudo-Huber loss $v = \sqrt{||\phi_\theta(x)||^2 + 1} - 1$, which was also used in the experiments.

Owing to the link function, HSC has a natural probability estimate of $\hat{\eta} = 1 - e^{-v}$ from which the logit is computed with $\sigma^{-1}$. Either $v$ or $1 - e^{-v}$ can be used as the anomaly score, however for the experiments the former is chosen for improved numerical stability over the latter.

Note that HSC is not a proper composite loss (see proposition 5 in Appendix A), however it does make use of the strictly proper CPE log loss.

### 4.1.3  Fully Convolutional Data Description (FCDD)

FCDD loss [34] is directly inspired by the HSC loss, and can be seen as a generalization to fully convolutional networks $\phi_\theta : \mathbb{R}^{c \times h \times w} \to \mathbb{R}^{m \times n}$ that reduce the dimensions of an image $x \in \mathbb{R}^{c \times h \times w}$ of height and width $(h, w) < (m, n)$, with $c$ channels. An anomaly score for each pixel of $\phi_\theta$ output indexed by $(i, j)$ is computed using the element-wise pseudo-Huber loss to produce an anomaly heatmap: $A_{ij} = \sqrt{(\phi_\theta(x)^2)_{ij} + 1} - 1$. The loss is then computed as for HSC

$$\lambda(y, v) = -y \ln(1 - e^{-v}) + (1 - y)v$$

with $v = \frac{1}{mn} \sum_{i,j} A_{ij}$ being the average of the per-pixel anomaly scores in $A$.

This loss is used only for localization experiments, which requires an anomaly heatmap at the size of the original image both for inference, and per-pixel post-hoc calibration. To achieve this, at test time FCDD upsamples $A$ to $A' \in \mathbb{R}^{h \times w}$ through transpose convolution with a fixed Gaussian filter that has kernel size, padding, and stride determined by the effective receptive fields of pixels in $A$ with respect to input $x$. The filter's standard deviation was chosen as 14 in the experiments.

For pixel $(i, j)$, the per-pixel anomaly score is $A'_{ij}$, probability estimate is $\hat{\eta} = 1 - e^{-A'_{ij}}$, and logit is $z = \sigma^{-1}(\hat{\eta})$.

#### 4.1.4 Structural Similarity Index Measure (SSIM)

SSIM [49] is a metric often used in computer vision to test the similarity between two images $P$ and $Q$ of the same size. It's applied to the images locally for a pair of patches $p$ and $q$ of the same size and location in their respective images, and computed as

$$\text{SSIM}(p, q) = \frac{(2\mu_p\mu_q + c_1)(2\sigma_{pq} + c_2)}{(\mu_p^2 + \mu_q^2 + c_1)(\sigma_p^2 + \sigma_q^2 + c_2)} \in [-1, 1]$$

where $\mu_p$ is the mean of $p$, $\sigma_p$ the standard deviation of $p$, and $\sigma_{pq}$ the covariance between $p$ and $q$. Perfect similarity is given as 1 for two identical patches $p = q$.

A sliding window can be used to selected $p, q$ patches from a pair of images to produce an SSIM matrix $S(P, Q)$. This allows for training a reconstruction-based autoencoder $\phi_\theta$ from a derived unsupervised loss $\ell(x)$, as in [5], such that optimization encourages improvement in the similarity between images. Denoting $\mathbf{1}_{h \times w}$ as an $h \times w$-sized matrix with all elements equal to 1, for input $x$ as an $h \times w$ input image the loss is given as

$$\ell(x) = \frac{1}{hw} \sum_{i,j} (\mathbf{1}_{h \times w} - S(x, \phi_\theta(x)))_{ij}$$

where $i, j$ are pixel indices.

Since all elements of $S$ are bounded and in $[-1, 1]$, for localization a natural choice for the per-pixel estimate is $\hat{\eta}_{ij} = (1 - S_{ij}(x, \phi_\theta(x)))/2$, with corresponding logit $z_{ij} = \sigma^{-1}(\hat{\eta}_{ij})$ and anomaly score $\text{AS} = 2\hat{\eta}$. Detection averages $\hat{\eta}_{ij}$, from which the logit and anomaly score are derived.

For all experiments, a window size of $11 \times 11$ was used for computing $S$ after constant padding of $x$ and $\phi_\theta(x)$, by 5 pixels with the mean of the training data, such that $S$ has conformal width and height for comparison against segmentation masks.

## 4.2 Experimental Methodology

### 4.2.1 Training

Experiments are conducted using data from a class as normal data, which is split into training and calibration sets in a 3:1 ratio. 5 neural networks were trained for 5 random seeds on each base loss outlined in section 4.1 using the training set, with supervised losses including outlier exposure (OE) data from auxiliary datasets as specified below. For anomaly localization, each pixel of the normal data is labeled with normal class 0, and each of the pixels of synthetic anomalous data with anomalous class 1. Each network was then post-hoc calibrated with Platt scaling or Beta calibration, optimized over the logistic loss for anomaly detection and average per-pixel logistic loss for anomaly localization, using either outlier exposure or random, spectral images, generated as described in section 3.4, as synthetic anomalous data. Experiments for the calibration head method is also conducted for anomaly detection, but not localization. For comparison of these post-hoc calibrated models, a network was trained on the base loss using the full training data. If outlier exposure is used for both training and calibration, care is taken to ensure the sets for both phases are disjoint.

Networks were optimized for both training and calibration using ADAM [26] with initial learning rate of $10^{-4}$ and no weight decay, except for post-hoc Platt scaling and $\beta$ calibration for anomaly detection which used the Limited-memory BFGS optimizer [32] using $10,000$ datapoints each from augmented normal and synthetic anomalous data. ADAM optimization was performed over a batch size of 128 images, with each batch balanced to include an equal number of normal and synthetic anomalous data for supervised losses. Learning rate scheduling schemes and number of epochs for ADAM on each dataset are further outlined below.

Image augmentation was applied to normal class and synthetic anomalous outlier exposure data with color jitter, addition of small Gaussian noise, and flipping about the vertical axis (except for the classes of MVTecAD that exhibit bias against vertical symmetry in the test data). After augmentation, for the SSIM loss images were converted to grayscale, and in all cases data was normalized to $z$-scores of the standard normal distribution using the mean and standard deviation of the training split or full training data, depending on whether the network is to be post-hoc calibrated or not respectively.

Network architectures and other dataset-specific details are provided below:

- Fashion MNIST [50] - The LeNet [30]-style network as used in [35] was trained over 200 epochs with learning rate reduction by a factor of 10 at epoch milestones 100 and 150. For SSIM loss, a convolutional encoder was constructed by decapitating the fully connected layers, and appending a convolutional layer with output to 100 latent features in the bottleneck. The decoder was constructed by reflection of the encoder, replacing convolutions with transpose convolutions, producing the full autoencoder. The CIFAR-100 [27] was used for outlier exposure data where relevant.

- CIFAR-10 [27] - The LeNet [30]-style network as used in [35] was trained over 200 epochs with learning rate reduction by a factor of 10 at epoch milestones 100 and 150. For SSIM loss, a convolutional encoder was constructed by decapitating the fully connected layers and appending two convolutional layers, with output of the first to dimension $64 \times 8 \times 8$ and of the second to 100 latent features in the bottleneck. The decoder was constructed by reflection of the encoder, replacing convolutions with transpose convolutions, producing the full autoencoder. The CIFAR-100 [27] was used for outlier exposure data where relevant.

- MVTecAD [4] and MPDD [25] - Optimization was performed for 300 epochs with learning rate reduction by a factor of 10 at epoch milestones 200 and 250. Following [35], ImageNet21k with ImageNet1k data removed [15] was used for providing outlier exposure data. These datasets require image resizing for compatibility with the networks, which is done by first resizing the full image, then randomly cropping – if an input size of $256 \times 256$ is required, images are first resized to $292 \times 292$, whereas if a network requires $224 \times 224$ input, images are first resized to $256 \times 256$. Both of these datasets were used for both anomaly detection and localization, with architecture details for each task specified below.

    - Anomaly detection - A WideResNet [54] network with ResNet-18 backbone as used in [22] was used for all losses except SSIM loss, which used the autoencoder

for objects of MVTecAD as outlined in [4]. The former required images to be resized to $224 \times 224$, and the latter to $256 \times 256$.

 - Anomaly localization - The same autoencoder as for the detection task was used for SSIM loss, from which a U-Net style network [41] was constructed for use by average pixel-wise logistic loss by adding skip connections. For FCDD, the same network in [34], that employs a decapitated, pretrained VGG11 network with batch normalization as backbone, was used. The autoencoder and U-Net networks required images to be resized to $256 \times 256$, and the FCDD network to $224 \times 224$.

### 4.2.2 Evaluation

Test data was normalized to the $z$-score of the standard normal distribution using the training data mean and standard deviation, grayscaled for networks trained on SSIM loss, and if necessary resized without cropping to the relevant input size of the network.

For anomaly detection, Fashion MNIST and CIFAR-10 are evaluated in a one-vs-all manner due to the lack of representative anomalous data for a given class. Test data for the normal class is used as normal test data, and the complement of the normal class over all test data is used as anomalous data. MVTecAD and MPDD datasets contain normal and anomalous data for each class, hence detection was evaluated for each class in isolation.

For evaluating calibration, careful data consideration must be taken for computing the ECE and MCE as described in section 2.4.3. Since the models are calibrated against synthetic anomalous data, they will be badly calibrated against the anomalous data used for testing. For that reason, we evaluate calibration against the normal test data and an equal-sized set of the synthetic anomalies that weren't used for either training or post-hoc calibration.

For detection, AUROC, ECE and MCE are reported, and for localization the AUPRO and per-pixel AUROC, ECE and MCE. AUROC and AUPRO are also reported for input-perturbed test data using $\epsilon = 1.4 \times 10^{-3}$ as outlined in section 3.3.

## 4.3 Results

### 4.3.1 Anomaly Detection

Average results over all classes for the one-vs-all detection experiments can be found in table 4.1, and for per-class-detection experiments in table 4.2, while results for each class can be found in Appendix B.

Input perturbation is shown to be an effective technique for post-hoc calibrated models, resulting in AUROC improvements across all of them. While it is also effective for base models fully trained over the supervised logistic and HSC losses, the same can't be said for the unsupervised SVDD and SSIM losses, where perturbation causes only marginal improvements for the one-vs-all datasets, while reducing the AUROC for per-class-detection datasets.

For the one-vs-all experiments, base networks fully trained on supervised base losses yield the best AUROC with input perturbation, while for per-class-detection the best results with perturbation are provided by their post-hoc calibration head variants. This suggests that supervised losses can benefit from post-hoc calibration in data settings where anomalies are subtle and local in the images, rather than global as induced by a one-vs-all setting. On the other hand, it is also plausible that the relatively small number of training samples in the per-class-detection datasets compared to the one-vs-all datasets plays a role.

| | Loss | Metric | Fully Trained | CalHead OE | CalHead Spectral | Platt OE | Platt Spectral | $\beta$ OE | $\beta$ Spectral |
|---|---|---|---|---|---|---|---|---|---|
| Fashion MNIST | SVDD | AUROC | 72.53 | 74.73 | 71.63 | 73.80 | 73.80 | 73.47 | 73.80 |
| | | | 72.67 | **78.08** | 75.13 | 77.92 | 77.92 | 77.52 | 77.92 |
| | | MCE | **1.79** | 26.03 | 25.18 | 24.84 | 27.80 | 22.87 | 23.88 |
| | | ECE | 0.43 | 0.63 | **0.40** | 1.39 | 1.85 | 0.81 | 0.70 |
| | SSIM | AUROC | 85.90 | 89.03 | 82.06 | 85.74 | 85.74 | 85.74 | 85.74 |
| | | | 85.96 | **90.50** | 84.34 | 86.89 | 86.89 | 86.89 | 86.89 |
| | | MCE | 73.10 | 11.26 | **1.04** | 17.46 | 14.87 | 15.03 | 14.02 |
| | | ECE | 9.14 | 0.09 | **0.07** | 0.17 | 0.19 | 0.16 | 0.17 |
| | LGS | AUROC | 85.66 | 85.46 | 85.71 | 85.44 | 85.44 | 75.38 | 75.38 |
| | | | **88.60** | 88.32 | 88.46 | 88.19 | 87.90 | 75.48 | 75.46 |
| | | MCE | 0.03 | 0.03 | **0.02** | 0.03 | 0.02 | 0.04 | 0.03 |
| | | ECE | 0.01 | 0.01 | **0.01** | 0.01 | 0.01 | 0.01 | 0.01 |
| | HSC | AUROC | 83.96 | 74.82 | 72.49 | 78.57 | 78.57 | 78.57 | 78.57 |
| | | | **91.36** | 81.17 | 78.63 | 79.23 | 79.18 | 79.42 | 79.40 |
| | | MCE | 0.03 | 0.18 | 0.83 | 0.03 | 0.02 | 0.03 | **0.02** |
| | | ECE | 0.01 | 0.05 | 0.10 | 0.01 | 0.01 | 0.01 | **0.01** |
| CIFAR-10 | SVDD | AUROC | 52.97 | 69.22 | 60.12 | 53.22 | 53.22 | 53.22 | 53.22 |
| | | | 53.29 | **79.02** | 70.21 | 62.82 | 62.82 | 62.82 | 62.82 |
| | | MCE | **1.60** | 14.35 | 18.51 | 10.39 | 21.43 | 10.37 | 17.38 |
| | | ECE | **0.19** | 0.94 | 0.74 | 0.81 | 1.52 | 0.81 | 1.16 |
| | SSIM | AUROC | 58.97 | 75.95 | 67.84 | 59.21 | 59.21 | 59.21 | 59.21 |
| | | | 58.99 | **78.79** | 71.69 | 64.23 | 64.23 | 64.23 | 64.23 |
| | | MCE | 70.42 | **7.89** | 8.70 | 16.26 | 11.81 | 14.16 | 18.27 |
| | | ECE | 14.92 | 0.27 | **0.27** | 0.65 | 2.73 | 0.78 | 2.94 |
| | LGS | AUROC | 96.53 | 96.33 | 90.68 | 96.13 | 96.13 | 89.40 | 89.40 |
| | | | **98.42** | 98.18 | 94.41 | 98.10 | 98.10 | 89.63 | 89.63 |
| | | MCE | 69.20 | 15.38 | **12.05** | 16.63 | 14.66 | 18.52 | 17.00 |
| | | ECE | 1.06 | 0.18 | **0.09** | 0.22 | 0.11 | 0.79 | 1.41 |
| | HSC | AUROC | 96.47 | 95.97 | 84.50 | 95.75 | 95.75 | 95.75 | 95.75 |
| | | | **98.36** | 98.04 | 90.52 | 97.58 | 97.57 | 97.58 | 97.57 |
| | | MCE | 64.35 | 16.45 | 11.46 | 17.29 | 9.00 | 16.16 | **8.21** |
| | | ECE | 0.76 | 0.23 | 0.13 | 0.19 | 0.14 | 0.18 | **0.12** |

Table 4.1: % AUROC, MCE and ECE for SVDD, SSIM, logistic (LGS) and HSC losses in the one-vs-all experiments on Fashion MNIST and CIFAR-10 datasets. AUROC over unperturbed and perturbed test inputs is shown in the white and gray rows respectively, with the largest value per loss emphasized in bold font. The smallest MCE and ECE per loss are also emphasized in bold font.

| | Loss | Metric | Fully Trained | CalHead OE | CalHead Spectral | Platt OE | Platt Spectral | β OE | β Spectral |
|---|---|---|---|---|---|---|---|---|---|
| **MVTecAD** | SVDD | AUROC | 64.89 | 53.00 | 54.01 | 65.97 | 65.97 | 65.97 | 65.97 |
| | | | 62.68 | 92.82 | 92.47 | 97.62 | **97.67** | 97.62 | 97.67 |
| | | MCE | 41.00 | 52.05 | 46.91 | 7.88 | 2.51 | 9.59 | **2.03** |
| | | ECE | 7.65 | 5.09 | 6.74 | 0.14 | **0.12** | 0.18 | 0.18 |
| | SSIM | AUROC | 57.99 | 56.80 | 53.45 | 63.03 | 63.03 | 63.03 | 63.03 |
| | | | 58.33 | 74.99 | 72.28 | 82.71 | 82.96 | 82.94 | **82.97** |
| | | MCE | 57.05 | 47.72 | 43.96 | **20.10** | 21.06 | 20.38 | 20.39 |
| | | ECE | 6.35 | 1.81 | 1.73 | 1.43 | 1.47 | **1.32** | 1.54 |
| | LGS | AUROC | 63.90 | 62.65 | 59.86 | 63.63 | 63.51 | 56.31 | 55.98 |
| | | | 83.36 | **87.39** | 79.03 | 85.97 | 87.01 | 57.30 | 57.10 |
| | | MCE | 0.67 | 2.84 | 9.02 | 0.00 | **0.00** | 0.04 | 0.00 |
| | | ECE | 0.66 | 0.17 | 0.26 | 0.00 | **0.00** | 0.01 | 0.00 |
| | HSC | AUROC | 63.42 | 59.61 | 54.71 | 62.79 | 62.79 | 62.79 | 62.79 |
| | | | 71.14 | **95.83** | 81.87 | 75.21 | 76.11 | 76.18 | 76.24 |
| | | MCE | 0.03 | 44.93 | 42.47 | 0.17 | **0.00** | 0.34 | 0.00 |
| | | ECE | 0.01 | 11.03 | 7.55 | 0.00 | **0.00** | 0.02 | 0.00 |
| **MPDD** | SVDD | AUROC | 69.02 | 51.75 | 60.69 | 72.61 | 72.61 | 72.61 | 72.61 |
| | | | 68.85 | 67.67 | 79.52 | 89.14 | **89.17** | 89.13 | 89.13 |
| | | MCE | 42.93 | 24.90 | 37.81 | 0.62 | 0.35 | 0.57 | **0.32** |
| | | ECE | 16.28 | 9.87 | 12.53 | 0.16 | **0.08** | 0.15 | 0.08 |
| | SSIM | AUROC | 67.87 | 61.63 | 57.15 | 71.70 | 71.70 | 71.70 | 71.70 |
| | | | 66.93 | 73.28 | 68.98 | 84.26 | 84.08 | 84.26 | **84.26** |
| | | MCE | 63.95 | 46.40 | 47.32 | 9.36 | 7.30 | 9.14 | **6.98** |
| | | ECE | 8.13 | 2.00 | 2.24 | 0.32 | 0.21 | 0.29 | **0.20** |
| | LGS | AUROC | 58.49 | 58.53 | 55.61 | 58.29 | 58.29 | 59.16 | 59.16 |
| | | | 71.04 | **71.75** | 66.02 | 69.58 | 70.77 | 60.28 | 60.28 |
| | | MCE | **0.11** | 0.81 | 1.71 | 0.40 | 0.31 | 0.40 | 0.20 |
| | | ECE | **0.03** | 0.28 | 0.49 | 0.10 | 0.08 | 0.10 | 0.05 |
| | HSC | AUROC | 68.82 | 63.30 | 63.00 | 71.21 | 71.21 | 71.21 | 71.21 |
| | | | 76.70 | **96.19** | 81.84 | 80.03 | 81.41 | 81.38 | 81.48 |
| | | MCE | 0.05 | 46.77 | 47.19 | 0.10 | **0.01** | 0.10 | 0.01 |
| | | ECE | 0.02 | 11.87 | 11.25 | 0.03 | **0.00** | 0.03 | 0.00 |

Table 4.2: % AUROC, MCE and ECE for SVDD, SSIM, logistic (LGS) and HSC losses in the per-class-detection experiments on MVTecAD and MPDD datasets. AUROC over unperturbed and perturbed test inputs is shown in the white and gray rows respectively, with the largest value per loss emphasized in bold font. The smallest MCE and ECE per loss are also emphasized in bold font.

Networks trained on unsupervised losses almost always see an improved AUROC from post-hoc calibration, with the largest improvements on the one-vs-all datasets resulting from the calibration head method, and for per-class-detection datasets from Platt scaling or $\beta$ calibration. In the latter case, synthetic anomalies from random spectral data always outperforms synthetic anomalies from outlier exposure datasets.

More generally, spectral data is as competitive as outlier exposure data for Platt scaling and $\beta$ calibration for improving detection results, oftentimes performing slightly better. However, spectral data is ineffective compared to outlier exposure data for the calibration head method. This can be attributed to the higher-dimensional input features, resulting from the outputs of a network's frozen layers, used for optimization of the calibration head, compared to the one-dimensional inputs, derived from a fully-frozen network's output, used for optimization of the Platt and $\beta$ models. In other words, this is a manifestation of the phenomenon described in section 3.4, where randomly generated synthetic anomalies result in poor performance for high-dimensional inputs.

Non-perturbed post-hoc calibrated methods tend to produce worse or the same result as their fully trained base counterpart. This makes sense, as a secondary post-hoc training phase for calibration deprives the initial training phase of data. We note improvements where this isn't true only occurs for the unsupervised losses in the one-vs-all experiments.

At a glance, the calibration metrics seem to have no relationship on perturbation-based AUROC improvements. To more rigorously investigate whether there is any relationship between either ECE or MCE of a model with the performance increase from input perturbation, their Spearman rank correlation coefficients were computed for each dataset. The performance increase $\kappa$ was computed as the difference between the perturbed result $\text{AUROC}_p$ and unperturbed result $\text{AUROC}_0$, relative to the distance from a perfect score to $\text{AUROC}_0$: $\kappa = (\text{AUROC}_p - \text{AUROC}_0)/(1 - \text{AUROC}_0)$. Overall, there were as many $(\kappa, \text{ECE})$ and $(\kappa, \text{MCE})$ pairs as there were models trained per dataset; i.e. with 5 seeds for each of $n_c$ number of classes over 7 methods tested for the 4 losses, there were $n = 140 \times n_c$ points to compute the correlation for a given dataset. Correlation coefficient magnitudes were between 0.03 and 0.45 for MCE and $10^{-4}$ and 0.185 for ECE, suggesting that there is no relationship between the calibration metrics and performance increase from input perturbation.

Degenerate optimization from numerical instability can be explored as one reason why this would be the case. Consider $\beta$ calibration for logistic loss, which frequently underperforms compared to Platt scaling. In most cases, Platt and $\beta$ optimization lead to the same results, as they are incredibly similar techniques and in the experiments are optimized over the same initially-trained base networks. However, as discussed in section 3.2, base networks trained on strictly proper losses lead to overconfidence in results, corresponding to very large magnitudes in logit outputs. This causes many of the calibration set's inputs to get mapped to an estimate $\hat{\eta}$ of 0 or 1 used for $\beta$ optimization, due to the lack of floating point precision required to fully represent these estimates derived from sigmoid evaluation of very large logits. The original logit rankings get destroyed by the sigmoid, so the $\beta$ parameters con-verge to a suboptimal solution, which empirically also impedes its ability to have improved AUROC results from test input perturbation. Getting back to the point, one can observe that $\beta$-calibrated models for logistic loss still produce about the same ECE and MCE as Platt-calibrated models, despite this numerical instability that's in plain sight from the poor AUROC results.

### 4.3.2 Anomaly Localization

Average results over all classes for the anomaly localization experiments can be found in table 4.3, while results for each class can be found in Appendix C.

Per-pixel post-hoc Platt scaling and $\beta$ calibration for localization are much less effective than for anomaly detection. For FCDD in particular, AUROC and AUPRO values without perturbation drastically decrease with post-hoc calibration to the point that the improvements from input perturbation don't result in an improvement against even the corresponding unperturbed fully-trained base FCDD values. In fact, FCDD results for MPDD shows that input perturbation is most effective for the fully-trained FCDD networks to begin with.

The perturbed AUROC and AUPRO in general provide marginal increases over unperturbed values in stark contrast to detection. Input perturbation for post-hoc calibrated models can still lead to improvements over the fully-trained base networks, as for SSIM on MVTecAD and the logistic loss on MPDD, however these are the exception to the rule. The main trend is that even in cases where a larger increase in these values for post-hoc calibrated models with perturbation is observed compared to the perturbation increase for the fully trained base networks, per-pixel post-hoc calibration harms localization performance to where the deficit can't be overcome.

In any case, test input perturbation does yield improved localization results over corresponding unperturbed experiments in all cases, where similarly to detection the smallest improvements occur for the unsupervised SSIM loss. In contrast to the detection results, post-hoc calibration of SSIM does not ameliorate this marginal improvement by perturbation, oftentimes maintaining the same increases.

One can also observe that for the logistic loss, the fully-trained networks are *more calibrated* than the post-hoc calibrated models. That is to say, post-hoc calibration may not even improve the calibration of the model.

Calibrating each output pixel of the anomaly map with Platt scaling or $\beta$ calibration is simply ineffective, most likely due the inherent location-dependence of these techniques. During inference, since the same strictly monotonic transformation is applied for a heatmap pixel, the post-hoc calibrated model does not adapt to the changing values of the network output over different images which are supposed to reflect the changing locations of anomalous pixels. In that sense, post-hoc calibration using these methods seems to be an exercise in finding some random permutation of the rankings of pixel-wise anomaly scores. Especially given that the model never actually sees representative anomalies during training or post-hoc calibration, meaning the per-pixel Platt or $\beta$ transformations can't even develop a representative bias towards anomaly locations for MVTecAD or MPDD that might ameliorate localization results.

| | Loss | Metric | Fully Trained | Platt OE | Platt Spectral | $\beta$ OE | $\beta$ Spectral |
|---|---|---|---|---|---|---|---|
| **MVTecAD** | FCDD | AUROC | 77.71 | 64.75 | 65.12 | 64.57 | 64.67 |
| | | | **79.50** | 66.70 | 67.08 | 66.62 | 66.77 |
| | | AUPRO | 48.57 | 30.73 | 30.92 | 30.36 | 30.82 |
| | | | **51.22** | 33.27 | 33.55 | 33.16 | 33.75 |
| | | MCE | 84.11 | 83.75 | 83.53 | 83.47 | **83.21** |
| | | ECE | 5.34 | 5.14 | 5.21 | **4.31** | 4.38 |
| | SSIM | AUROC | 67.36 | 66.80 | 66.91 | 66.69 | 66.91 |
| | | | 67.80 | 68.53 | **68.59** | 68.38 | 68.58 |
| | | AUPRO | 33.86 | 32.96 | 32.80 | 32.93 | 33.24 |
| | | | 34.65 | 34.61 | 34.44 | 34.56 | **34.87** |
| | | MCE | 50.63 | 48.35 | **46.71** | 49.33 | 46.99 |
| | | ECE | 1.79 | 1.45 | 1.44 | **1.23** | 1.32 |
| | LGS | AUROC | 64.79 | 61.92 | 61.59 | 59.66 | 59.64 |
| | | | **67.30** | 64.41 | 64.09 | 61.53 | 61.44 |
| | | AUPRO | 29.26 | 27.20 | 26.78 | 24.24 | 23.92 |
| | | | **31.53** | 29.33 | 29.03 | 26.19 | 25.88 |
| | | MCE | **14.53** | 19.77 | 17.58 | 22.51 | 16.09 |
| | | ECE | 0.01 | 0.00 | **0.00** | 0.01 | 0.01 |
| **MPDD** | FCDD | AUROC | 80.80 | 57.25 | 55.08 | 56.71 | 57.15 |
| | | | **83.70** | 57.52 | 55.57 | 57.00 | 57.47 |
| | | AUPRO | 56.08 | 21.33 | 19.72 | 21.48 | 22.22 |
| | | | **60.60** | 21.90 | 20.67 | 22.08 | 22.89 |
| | | MCE | **84.60** | 84.83 | 84.84 | 84.79 | 84.81 |
| | | ECE | 5.12 | 5.62 | 5.82 | **5.04** | 5.11 |
| | SSIM | AUROC | 86.99 | 85.89 | 85.98 | 85.03 | 85.13 |
| | | | **87.15** | 87.02 | 87.11 | 86.20 | 86.28 |
| | | AUPRO | 63.08 | 60.84 | 61.18 | 59.34 | 59.67 |
| | | | **63.54** | 62.10 | 62.42 | 60.65 | 60.96 |
| | | MCE | 45.48 | 39.81 | 43.28 | **37.24** | 39.31 |
| | | ECE | 2.16 | 1.92 | 1.97 | **1.72** | 1.77 |
| | LGS | AUROC | 43.09 | 44.44 | 45.15 | 54.98 | 55.01 |
| | | | 45.36 | 46.59 | 47.24 | 56.32 | **56.33** |
| | | AUPRO | 19.07 | 18.42 | 18.92 | 20.95 | 21.12 |
| | | | 20.90 | 20.14 | 20.63 | 22.44 | **22.54** |
| | | MCE | **28.20** | 31.59 | 33.82 | 36.68 | 36.44 |
| | | ECE | **0.02** | 0.02 | 0.02 | 0.03 | 0.02 |

Table 4.3: % AUPRO and pixel-wise AUROC, MCE and ECE for FCDD, SSIM and logistic (LGS) losses over the MVTecAD and MPDD datasets. AUPRO and Pixel-wise AUROC over unperturbed and perturbed test inputs are shown in the white and gray rows respectively, with the largest values per loss emphasized in bold font. The smallest pixel-wise MCE and ECE per loss are also emphasized in bold font.

# Chapter 5

# Conclusion

Post-hoc calibration is an effective technique for anomaly detection, especially for unsupervised losses for which calibration is not actively encouraged. In particular, these miscalibrated models become better calibrated through training on a strictly proper loss with a decreased number of parameters in the calibration phase, which directly leads to observed improvements in detection performance.

When paired with gradient-based input perturbation, detection results can see drastic improvements in post-hoc models compared to fully-trained baselines. This improvement is most prominent for datasets with a small number of training samples and local, region-level anomalies in the test set, whereby input perturbation is drastically more effective for post-hoc calibrated models over fully trained base models for every loss. On the other hand, it is still the case for larger datasets with global, image-level anomalies that fully training with strictly proper losses leads to the best results.

Platt scaling and $\beta$ calibration were found to produce similar results, save for when $\beta$ calibration suffers from numerical instability of overconfident networks trained on the logistic loss. As they both lead to strictly monotonic transformations, they are unable to improve results from the initially-trained networks. It makes it all-the-more striking the extent to which detection results are improved when paired with gradient-based perturbation, oftentimes outperforming the calibration head method that can lead to ranking permutations without perturbation.

Spectral data is also shown to be as effective as, and at times more effective than, outlier exposure as a source of synthetic anomalies for post-hoc calibration. This is hypothesized to occur since calibration induces a much smaller input space for which parameters are optimized during the calibration phase, which allows for traditional success with randomly generated data to be enjoyed, which suggests that the same observations could be made for using other forms of randomly generated data with sufficient diversity.

When applied to the localization task, post-hoc calibration was found to produce poor results. This can be explained by the inherent pixel location dependence of the per-pixel Platt scaling and $\beta$ calibration methods used. This flaw suggests that suitably-designed post-hoc calibration schemes that facilitate generalization to adaptive location changes of anomalous regions could improve localization, leaving the door open for future work. Regrettably, this work failed to provide any conclusive answer on whether concepts of post-hoc calibration can improve anomaly localization.

# Appendix A

# Proofs

Contained here are simple proofs for propositions referenced throughout the body of the thesis.

**PROPOSITION 2.** Let $\mathbb{P}$ be a joint distribution over $\mathcal{D} = \mathcal{X} \times \mathcal{Y}$ with tuple $Y = (Y_1, Y_2) \in \mathcal{Y} \subset \mathbb{N}^2$, and conditional distribution $\gamma := \mathbb{P}_{Y|X}(y|x)$. Assume $Y_1$ and $Y_2$ are conditionally independent given $X$ such that element-wise conditional distributions are given by $\gamma_1 = \mathbb{P}_{Y_1|X}(y_1|x)$ and $\gamma_2 = \mathbb{P}_{Y_2|X}(y_2|x)$. Then, the sum $\ell(y, \hat{\gamma})$ of two element-wise (strictly) proper losses $\ell_1(y_1, \hat{\gamma}_1)$ and $\ell_2(y_2, \hat{\gamma}_2)$ is (strictly) proper.

*Proof.* Conditional risk $L$ is bounded below by the sum of Bayes conditional risks $L_1^*(\hat{\gamma}_1)$ and $L_2^*(\hat{\gamma}_2)$:

$$
\begin{aligned}
L(\gamma, \hat{\gamma}) &= \underset{Y \sim \gamma}{\mathbb{E}} \left[ \ell_1(Y_1, \hat{\gamma}_1) + \ell_2(Y_2, \hat{\gamma}_2) \right] \\
&= \underset{Y_1 \sim \gamma_1}{\mathbb{E}} \left[ \ell_1(Y_1, \hat{\gamma}_1) \right] + \underset{Y_2 \sim \gamma_2}{\mathbb{E}} \left[ \ell_2(Y_2, \hat{\gamma}_2) \right] \\
&= L_1(\gamma_1, \hat{\gamma}_1) + L_2(\gamma_2, \hat{\gamma}_2) \\
&\leq L_1^*(\gamma_1) + L_2^*(\gamma_2)
\end{aligned}
$$

The second equality follows from the linearity of expectation and $\ell_i$ not being a function of $y_{j \neq i}$, and owing to propriety of $\ell_1$ and $\ell_2$, the infimum on the right of the inequality is achieved for element-wise minimizers $\hat{\gamma}_1 = \gamma_1$ and $\hat{\gamma}_2 = \gamma_2$.

Since $\gamma_1$ and $\gamma_2$ are conditionally independent given $X$, $\gamma = \gamma_1 \gamma_2$, implying that $L$ is minimized for $\hat{\gamma} = \gamma = \gamma_1 \gamma_2$, and hence $\ell$ is proper. If, furthermore, both element-wise losses are strictly proper, then the minimizer $\hat{\gamma} = \gamma_1 \gamma_2$ is unique since the element-wise minimizers $\gamma_1$ and $\gamma_2$ are unique, making $\ell$ strictly proper. ■

**PROPOSITION 3.** Log loss, given by equation 2.11, is strictly proper.

*Proof.* First, note that log loss is proper as it meets the stationarity condition of equation 2.10. The derivative of $\ell_1(\hat{\eta}) = -\ln(\hat{\eta})$ is $\ell_1'(\hat{\eta}) = -1/\hat{\eta}$, and of $\ell_0(\hat{\eta}) = -\ln(1 - \hat{\eta})$ is $\ell_0'(\hat{\eta}) = 1/(1 - \hat{\eta})$. Substituting into the stationarity condition shows that it's satisfied:

$$(1 - \eta)\ell_0'(\eta) = -\eta\ell_1'(\eta)$$
$$1 = 1$$

Since log loss is proper, it is minimized for $\hat{eta} = \eta$, and it is strictly proper as the second derivative of conditional risk is strictly positive, showing that its minimizer is unique:

$$\frac{\partial^2}{\partial\hat{\eta}^2}L(\eta,\hat{\eta}) = \frac{\partial^2}{\partial\hat{\eta}^2}(-\eta\ln(\hat{\eta}) - (1-\eta)\ln(1-\hat{\eta})) = \underbrace{\frac{\eta}{\hat{\eta}^2}}_{\geq 0} + \underbrace{\frac{1-\eta}{(1-\hat{\eta})^2}}_{\geq 0} > 0$$

The inequality holds since $\eta, \hat{\eta} \in [0,1]$, and if $\eta$ or $1-\eta$ is 0, the other is 1, implying that at least one of the additive terms is positive. ∎

**PROPOSITION 4.** The logistic loss, as given by equation 2.15, with logit link $\psi(\hat{\eta}) = \ln(\hat{\eta}) - \ln(1-\hat{\eta})$ and sigmoid inverse link $\psi^{-1}(v) = 1/(1+e^{-v})$, is strictly proper.

*Proof.* First, note that logistic loss is proper as it meets the stationarity condition for composite losses of equation 2.12. The derivative of $\lambda_1(v) = \ln(1+e^{-v})$ is $\lambda_1'(v) = -e^{-v}/(1+e^{-v}) = \psi^{-1}(v) - 1$, and of $\lambda_0(v) = \ln(1+e^v)$ is $\lambda_0'(v) = e^v/(1+e^v) = \psi^{-1}(v)$. Substituting into the stationarity condition shows that it's satisfied:

$$(1-\eta)\lambda_0'(\psi(\eta)) = -\eta\lambda_1'(\psi(\eta))$$
$$(1-\eta)\psi^{-1}(\psi(\eta)) = -\eta(\psi^{-1}(\psi(\eta)) - 1)$$
$$(1-\eta)\eta = (1-\eta)\eta$$

Since logistic loss is proper, it is minimized for $\hat{\eta} = \eta$, and it remains to show that this minimizer is unique. Making the substitution $\ell_i(\hat{\eta}) = \lambda_i(\psi(\hat{\eta}))$, the second derivative of conditional risk is given by:

$$\begin{aligned}
\frac{\partial^2}{\partial\hat{\eta}^2}L(\eta,\hat{\eta}) &= \frac{\partial^2}{\partial\hat{\eta}^2}(\eta\lambda_1(\psi(\hat{\eta})) + (1-\eta)\lambda_0(\psi(\hat{\eta}))) \\
&= \frac{\partial^2}{\partial\hat{\eta}^2}\left(\eta\ln\left(\frac{1}{\hat{\eta}}\right) + (1-\eta)\ln\left(\frac{1}{1-\hat{\eta}}\right)\right) \\
&= \underbrace{\frac{\eta}{\hat{\eta}^2}}_{\geq 0} + \underbrace{\frac{1-\eta}{(1-\hat{\eta})^2}}_{\geq 0} \\
&> 0
\end{aligned}$$

The inequality holds since $\eta, \hat{\eta} \in [0,1]$, and if $\eta$ or $1-\eta$ is 0, the other is 1, implying that at least one of the additive terms is positive. Since the second derivative of conditional risk is strictly positive, the minimizer is unique, making the logistic loss strictly proper. ∎

**PROPOSITION 5.** The HSC loss given by equation 4.1 is not proper.

*Proof.* The derivative of $\lambda_1(v) = -\ln(1 - e^{-v})$ is $\lambda_1'(v) = 1/(1 - e^{-v})$, and of $\lambda_0(v) = v$ is $\lambda_0'(v) = 1$. For link function $\psi(\hat{\eta}) = -\ln(1 - \hat{\eta})$, $\lambda_1'(\psi(\hat{\eta})) = -1/\hat{\eta}$. Substituting into the stationarity condition for composite losses 2.12 shows that it is not met:

$$
\begin{aligned}
(1 - \eta)\lambda_0'(\psi(\eta)) &\neq -\eta\lambda_1'(\psi(\eta)) \\
1 - \eta &\neq 1
\end{aligned}
$$

∎

# Appendix B

# Per-Class Detection Results

## B.1  Fashion MNIST

| Label | Fully Trained | CalHead OE | CalHead Spectral | Platt OE | Platt Spectral | $\beta$ OE | $\beta$ Spectral |
|---|---|---|---|---|---|---|---|
| ankle boot | 96.65 | 71.61 | 50.21 | 92.07 | 92.07 | 92.07 | 92.07 |
| | **96.74** | 74.63 | 53.03 | 93.52 | 93.52 | 93.52 | 93.52 |
| bag | 72.42 | 72.15 | 63.79 | 74.59 | 74.59 | 74.59 | 74.59 |
| | 73.02 | 76.25 | 68.74 | 79.29 | **79.29** | 79.29 | 79.29 |
| coat | 65.21 | 75.83 | 73.47 | 74.94 | 74.94 | 74.94 | 74.94 |
| | 65.69 | 78.80 | 76.70 | **80.37** | 80.37 | 80.37 | 80.37 |
| dress | 71.55 | 83.15 | 81.47 | 67.78 | 67.77 | 64.44 | 67.77 |
| | 71.70 | **86.56** | 83.83 | 72.16 | 72.16 | 68.20 | 72.16 |
| pullover | 66.04 | 61.04 | 63.23 | 64.08 | 64.08 | 64.08 | 64.08 |
| | 66.11 | 65.87 | 68.84 | **69.28** | 69.28 | 69.28 | 69.28 |
| sandal | 71.90 | 81.32 | 79.95 | 70.72 | 70.72 | 70.72 | 70.72 |
| | 72.05 | **82.42** | 80.97 | 74.77 | 74.77 | 74.77 | 74.77 |
| shirt | 60.13 | 52.51 | 58.26 | 64.06 | 64.06 | 64.06 | 64.06 |
| | 59.94 | 58.30 | 64.31 | 69.05 | **69.05** | 69.05 | 69.05 |
| sneaker | 84.35 | 92.94 | 90.30 | 84.90 | 84.90 | 84.90 | 84.90 |
| | 84.57 | **93.97** | 91.87 | 87.56 | 87.56 | 87.56 | 87.56 |
| top | 66.11 | 63.72 | 65.36 | 66.22 | 66.22 | 66.22 | 66.22 |
| | 66.54 | 70.25 | **71.77** | 71.60 | 71.60 | 71.60 | 71.60 |
| trouser | 70.91 | 92.99 | 90.22 | 78.67 | 78.67 | 78.67 | 78.67 |
| | 70.32 | **93.78** | 91.20 | 81.58 | 81.58 | 81.58 | 81.58 |
| Average | 72.53 | 74.73 | 71.63 | 73.80 | 73.80 | 73.47 | 73.80 |
| | 72.67 | **78.08** | 75.13 | 77.92 | 77.92 | 77.52 | 77.92 |

Table B.1: Average % AUROC over 5 seeds with the SVDD loss for each class of the Fashion MNIST dataset. AUROC over unperturbed and perturbed test inputs is shown in the white and gray rows respectively, with the largest value per class emphasized in bold font.

| Label | Fully Trained | CalHead OE | CalHead Spectral | Platt OE | Platt Spectral | $\beta$ OE | $\beta$ Spectral |
|---|---|---|---|---|---|---|---|
| ankle boot | 97.38 | 97.30 | 77.86 | 97.60 | 97.60 | 97.60 | 97.60 |
| | 97.39 | 97.77 | 79.76 | **97.86** | 97.86 | **97.86** | **97.86** |
| bag | 77.74 | 82.30 | 72.48 | 77.42 | 77.42 | 77.42 | 77.42 |
| | 77.81 | **84.67** | 75.99 | 79.49 | 79.49 | 79.49 | 79.49 |
| coat | 83.01 | 89.98 | 87.12 | 82.12 | 82.12 | 82.12 | 82.12 |
| | 83.06 | **91.46** | 89.08 | 83.86 | 83.86 | 83.86 | 83.86 |
| dress | 89.74 | 88.27 | 85.02 | 88.13 | 88.13 | 88.13 | 88.13 |
| | 89.71 | **89.74** | 87.22 | 89.00 | 89.00 | 89.00 | 89.00 |
| pullover | 80.30 | 84.57 | 78.66 | 82.05 | 82.05 | 82.05 | 82.05 |
| | 80.46 | **86.94** | 81.89 | 83.58 | 83.58 | 83.58 | 83.58 |
| sandal | 82.48 | 92.83 | 87.80 | 82.38 | 82.38 | 82.38 | 82.38 |
| | 82.59 | **93.84** | 89.33 | 83.52 | 83.52 | 83.52 | 83.52 |
| shirt | 72.06 | 76.98 | 72.52 | 71.37 | 71.37 | 71.37 | 71.37 |
| | 72.22 | **79.57** | 75.79 | 73.41 | 73.41 | 73.41 | 73.41 |
| sneaker | 96.34 | 94.77 | 88.47 | 94.43 | 94.43 | 94.43 | 94.43 |
| | **96.35** | 95.39 | 89.58 | 94.87 | 94.87 | 94.87 | 94.87 |
| top | 81.51 | 86.97 | 79.34 | 85.17 | 85.17 | 85.17 | 85.17 |
| | 81.53 | **88.71** | 82.23 | 86.29 | 86.29 | 86.29 | 86.29 |
| trouser | 98.44 | 96.33 | 91.34 | 96.74 | 96.74 | 96.74 | 96.74 |
| | **98.45** | 96.89 | 92.55 | 97.02 | 97.02 | 97.02 | 97.02 |
| Average | 85.90 | 89.03 | 82.06 | 85.74 | 85.74 | 85.74 | 85.74 |
| | 85.96 | **90.50** | 84.34 | 86.89 | 86.89 | 86.89 | 86.89 |

Table B.2: Average % AUROC over 5 seeds with the SSIM loss for each class of the Fashion MNIST dataset. AUROC over unperturbed and perturbed test inputs is shown in the white and gray rows respectively, with the largest value per class emphasized in bold font.

| Label | Fully Trained | CalHead OE | CalHead Spectral | Platt OE | Platt Spectral | $\beta$ OE | $\beta$ Spectral |
|---|---|---|---|---|---|---|---|
| ankle boot | 91.27 | 91.05 | 92.35 | 91.79 | 91.79 | 80.97 | 80.97 |
| | 93.04 | 93.11 | **94.08** | 93.69 | 93.47 | 81.09 | 81.05 |
| bag | 83.97 | 81.14 | 80.71 | 81.57 | 81.57 | 61.23 | 61.23 |
| | **87.34** | 84.39 | 83.94 | 84.81 | 84.03 | 61.44 | 61.34 |
| coat | 83.50 | 84.18 | 83.95 | 84.19 | 84.19 | 71.34 | 71.34 |
| | 86.78 | **87.00** | 86.64 | 86.95 | 86.50 | 71.38 | 71.36 |
| dress | 79.40 | 81.64 | 81.78 | 81.77 | 81.77 | 72.54 | 72.54 |
| | 83.39 | 85.44 | **85.48** | 85.30 | 85.25 | 72.58 | 72.58 |
| pullover | 89.66 | 88.97 | 88.65 | 88.96 | 88.96 | 73.63 | 73.63 |
| | **91.95** | 91.69 | 91.41 | 91.69 | 91.11 | 73.68 | 73.69 |
| sandal | 91.81 | 91.65 | 91.78 | 91.77 | 91.77 | 86.05 | 86.05 |
| | **93.78** | 93.63 | 93.66 | 93.51 | 93.72 | 86.12 | 86.12 |
| shirt | 76.61 | 76.05 | 74.64 | 75.14 | 75.14 | 68.78 | 68.78 |
| | **80.96** | 80.43 | 79.20 | 79.64 | 78.96 | 69.00 | 68.97 |
| sneaker | 94.20 | 94.89 | 95.58 | 94.65 | 94.65 | 89.94 | 89.94 |
| | 95.94 | 96.30 | **96.79** | 95.87 | 96.06 | 89.95 | 89.95 |
| top | 76.51 | 76.41 | 79.21 | 76.37 | 76.37 | 69.09 | 69.09 |
| | 81.08 | 80.64 | **82.89** | 80.38 | 79.78 | 69.30 | 69.30 |
| trouser | 89.64 | 88.60 | 88.47 | 88.19 | 88.19 | 80.26 | 80.26 |
| | **91.71** | 90.53 | 90.46 | 90.03 | 90.07 | 80.28 | 80.28 |
| Average | 85.66 | 85.46 | 85.71 | 85.44 | 85.44 | 75.38 | 75.38 |
| | **88.60** | 88.32 | 88.46 | 88.19 | 87.90 | 75.48 | 75.46 |

Table B.3: Average % AUROC over 5 seeds with the logistic loss for each class of the Fashion MNIST dataset. AUROC over unperturbed and perturbed test inputs is shown in the white and gray rows respectively, with the largest value per class emphasized in bold font.

| Label | Fully Trained | CalHead OE | CalHead Spectral | Platt OE | Platt Spectral | $\beta$ OE | $\beta$ Spectral |
|---|---|---|---|---|---|---|---|
| ankle boot | 92.89 | 80.57 | 77.54 | 82.82 | 82.82 | 82.82 | 82.82 |
| | **96.49** | 85.52 | 81.69 | 83.07 | 83.09 | 83.07 | 83.11 |
| bag | 72.27 | 57.91 | 49.80 | 63.70 | 63.70 | 63.70 | 63.70 |
| | **84.51** | 68.31 | 60.72 | 65.11 | 64.65 | 65.33 | 65.51 |
| coat | 79.67 | 67.92 | 68.25 | 75.47 | 75.47 | 75.47 | 75.47 |
| | **89.20** | 74.81 | 74.48 | 76.05 | 76.17 | 75.91 | 76.03 |
| dress | 78.72 | 73.46 | 72.92 | 76.13 | 76.13 | 76.13 | 76.13 |
| | **86.67** | 79.49 | 78.67 | 76.52 | 76.44 | 76.64 | 76.56 |
| pullover | 80.07 | 75.20 | 72.93 | 76.38 | 76.38 | 76.38 | 76.38 |
| | **90.81** | 82.05 | 79.91 | 78.05 | 77.56 | 78.22 | 77.66 |
| sandal | 90.28 | 85.00 | 83.66 | 87.17 | 87.17 | 87.17 | 87.17 |
| | **94.25** | 89.20 | 87.34 | 87.37 | 87.33 | 87.52 | 87.52 |
| shirt | 73.58 | 67.43 | 65.96 | 71.25 | 71.25 | 71.25 | 71.25 |
| | **85.06** | 75.66 | 73.13 | 72.42 | 73.17 | 73.98 | 73.98 |
| sneaker | 95.49 | 92.05 | 90.56 | 91.77 | 91.77 | 91.77 | 91.77 |
| | **97.73** | 94.38 | 92.57 | 91.84 | 91.82 | 91.84 | 91.84 |
| top | 80.32 | 66.59 | 61.16 | 71.56 | 71.56 | 71.56 | 71.56 |
| | **90.48** | 75.71 | 71.57 | 72.27 | 71.94 | 72.08 | 72.21 |
| trouser | 96.33 | 82.11 | 82.18 | 89.44 | 89.44 | 89.44 | 89.44 |
| | **98.36** | 86.55 | 86.18 | 89.60 | 89.61 | 89.65 | 89.62 |
| Average | 83.96 | 74.82 | 72.49 | 78.57 | 78.57 | 78.57 | 78.57 |
| | **91.36** | 81.17 | 78.63 | 79.23 | 79.18 | 79.42 | 79.40 |

Table B.4: Average % AUROC over 5 seeds with the HSC loss for each class of the Fashion MNIST dataset. AUROC over unperturbed and perturbed test inputs is shown in the white and gray rows respectively, with the largest value per class emphasized in bold font.

## B.2  CIFAR-10

| Label | Fully Trained | CalHead OE | CalHead Spectral | Platt OE | Platt Spectral | $\beta$ OE | $\beta$ Spectral |
|---|---|---|---|---|---|---|---|
| airplane | 55.06 | 69.46 | 59.34 | 48.50 | 48.50 | 48.50 | 48.50 |
|  | 55.59 | **79.34** | 66.43 | 56.47 | 56.47 | 56.47 | 56.47 |
| automobile | 51.85 | 74.03 | 57.98 | 50.15 | 50.15 | 50.15 | 50.15 |
|  | 52.04 | **83.18** | 68.42 | 62.05 | 62.05 | 62.05 | 62.05 |
| bird | 53.33 | 63.02 | 55.80 | 58.14 | 58.14 | 58.14 | 58.14 |
|  | 53.44 | **74.57** | 63.84 | 66.89 | 66.89 | 66.89 | 66.89 |
| cat | 50.51 | 63.72 | 54.30 | 51.98 | 51.98 | 51.98 | 51.98 |
|  | 50.31 | **76.86** | 68.81 | 63.05 | 63.05 | 63.05 | 63.05 |
| deer | 56.43 | 74.16 | 70.52 | 64.91 | 64.91 | 64.91 | 64.91 |
|  | 57.42 | **82.11** | 77.53 | 74.70 | 74.70 | 74.70 | 74.70 |
| dog | 51.02 | 65.16 | 53.87 | 51.98 | 51.98 | 51.98 | 51.98 |
|  | 51.28 | **75.93** | 66.46 | 61.84 | 61.84 | 61.84 | 61.84 |
| frog | 47.81 | 73.54 | 67.51 | 58.45 | 58.45 | 58.45 | 58.45 |
|  | 48.46 | **82.11** | 76.09 | 68.99 | 68.99 | 68.99 | 68.99 |
| horse | 55.69 | 63.18 | 55.98 | 49.07 | 49.07 | 49.07 | 49.07 |
|  | 56.15 | **75.30** | 70.04 | 58.80 | 58.80 | 58.80 | 58.80 |
| ship | 55.78 | 73.39 | 61.65 | 50.55 | 50.55 | 50.55 | 50.55 |
|  | 55.85 | **81.26** | 71.29 | 58.69 | 58.69 | 58.69 | 58.69 |
| truck | 52.18 | 72.59 | 64.21 | 48.44 | 48.44 | 48.44 | 48.44 |
|  | 52.39 | **79.56** | 73.24 | 56.75 | 56.75 | 56.75 | 56.75 |
| Average | 52.97 | 69.22 | 60.12 | 53.22 | 53.22 | 53.22 | 53.22 |
|  | 53.29 | **79.02** | 70.21 | 62.82 | 62.82 | 62.82 | 62.82 |

Table B.5: Average % AUROC over 5 seeds with the SVDD loss for each class of the CIFAR-10 dataset. AUROC over unperturbed and perturbed test inputs is shown in the white and gray rows respectively, with the largest value per class emphasized in bold font.

| Label | Fully Trained | CalHead OE | CalHead Spectral | Platt OE | Platt Spectral | $\beta$ OE | $\beta$ Spectral |
|---|---|---|---|---|---|---|---|
| airplane | 71.89 | 79.63 | 75.09 | 71.96 | 71.96 | 71.96 | 71.96 |
| | 72.12 | **82.66** | 78.67 | 76.16 | 76.16 | 76.16 | 76.16 |
| automobile | 58.32 | 79.68 | 75.83 | 59.56 | 59.56 | 59.56 | 59.56 |
| | 57.86 | **82.20** | 78.66 | 64.58 | 64.58 | 64.58 | 64.58 |
| bird | 60.13 | 71.01 | 62.45 | 60.78 | 60.78 | 60.78 | 60.78 |
| | 60.59 | **74.68** | 66.52 | 66.18 | 66.18 | 66.18 | 66.18 |
| cat | 59.90 | 66.33 | 54.59 | 60.06 | 60.06 | 60.06 | 60.06 |
| | 59.88 | **69.13** | 59.43 | 65.23 | 65.23 | 65.23 | 65.23 |
| deer | 49.84 | 74.42 | 66.76 | 49.11 | 49.11 | 49.11 | 49.11 |
| | 50.43 | **77.36** | 71.17 | 55.28 | 55.28 | 55.28 | 55.28 |
| dog | 70.22 | 72.59 | 56.18 | 68.99 | 68.99 | 68.99 | 68.99 |
| | 70.13 | **75.37** | 60.73 | 73.70 | 73.70 | 73.70 | 73.70 |
| frog | 35.79 | 75.62 | 69.89 | 38.08 | 38.08 | 38.08 | 38.08 |
| | 35.86 | **78.74** | 73.94 | 43.48 | 43.48 | 43.48 | 43.48 |
| horse | 59.65 | 75.44 | 64.08 | 58.66 | 58.66 | 58.66 | 58.66 |
| | 59.42 | **78.79** | 68.53 | 63.76 | 63.76 | 63.76 | 63.76 |
| ship | 73.05 | 83.24 | 75.86 | 72.85 | 72.85 | 72.85 | 72.85 |
| | 73.19 | **85.48** | 79.07 | 76.80 | 76.80 | 76.80 | 76.80 |
| truck | 50.86 | 81.57 | 77.62 | 52.02 | 52.02 | 52.02 | 52.02 |
| | 50.44 | **83.47** | 80.20 | 57.08 | 57.08 | 57.08 | 57.08 |
| Average | 58.97 | 75.95 | 67.84 | 59.21 | 59.21 | 59.21 | 59.21 |
| | 58.99 | **78.79** | 71.69 | 64.23 | 64.23 | 64.23 | 64.23 |

Table B.6: Average % AUROC over 5 seeds with the SSIM loss for each class of the CIFAR-10 dataset. AUROC over unperturbed and perturbed test inputs is shown in the white and gray rows respectively, with the largest value per class emphasized in bold font.

| Label | Fully Trained | CalHead OE | CalHead Spectral | Platt OE | Platt Spectral | $\beta$ OE | $\beta$ Spectral |
|---|---|---|---|---|---|---|---|
| airplane | 97.02 | 96.52 | 93.96 | 96.46 | 96.46 | 90.67 | 90.67 |
| | **98.41** | 98.02 | 96.41 | 98.03 | 98.03 | 90.84 | 90.83 |
| automobile | 99.01 | 98.91 | 97.61 | 98.87 | 98.87 | 95.12 | 95.12 |
| | **99.45** | 99.40 | 98.66 | 99.38 | 99.38 | 95.18 | 95.18 |
| bird | 93.94 | 93.36 | 81.56 | 93.10 | 93.10 | 84.52 | 84.52 |
| | **97.49** | 96.85 | 88.93 | 96.66 | 96.66 | 84.91 | 84.91 |
| cat | 91.11 | 91.15 | 82.90 | 90.48 | 90.48 | 80.03 | 80.03 |
| | **95.95** | 95.62 | 89.26 | 95.42 | 95.42 | 80.61 | 80.61 |
| deer | 96.86 | 96.52 | 87.43 | 96.43 | 96.43 | 89.31 | 89.31 |
| | **98.94** | 98.71 | 93.44 | 98.68 | 98.68 | 89.52 | 89.52 |
| dog | 94.83 | 94.67 | 89.16 | 94.29 | 94.29 | 86.57 | 86.57 |
| | **97.55** | 97.20 | 93.26 | 97.05 | 97.05 | 86.96 | 86.96 |
| frog | 98.19 | 98.08 | 89.01 | 97.82 | 97.82 | 91.16 | 91.16 |
| | **99.27** | 99.20 | 93.51 | 99.09 | 99.09 | 91.27 | 91.27 |
| horse | 98.46 | 98.32 | 96.90 | 98.31 | 98.31 | 92.63 | 92.63 |
| | **99.33** | 99.23 | 98.42 | 99.25 | 99.25 | 92.73 | 92.73 |
| ship | 98.36 | 98.29 | 96.38 | 98.24 | 98.24 | 93.57 | 93.57 |
| | **99.05** | 99.00 | 97.70 | 98.97 | 98.97 | 93.66 | 93.66 |
| truck | 97.56 | 97.46 | 91.88 | 97.29 | 97.29 | 90.40 | 90.40 |
| | **98.73** | 98.57 | 94.51 | 98.50 | 98.50 | 90.59 | 90.59 |
| Average | 96.53 | 96.33 | 90.68 | 96.13 | 96.13 | 89.40 | 89.40 |
| | **98.42** | 98.18 | 94.41 | 98.10 | 98.10 | 89.63 | 89.63 |

Table B.7: Average % AUROC over 5 seeds with the logistic loss for each class of the CIFAR-10 dataset. AUROC over unperturbed and perturbed test inputs is shown in the white and gray rows respectively, with the largest value per class emphasized in bold font.

| Label | Fully Trained | CalHead OE | CalHead Spectral | Platt OE | Platt Spectral | $\beta$ OE | $\beta$ Spectral |
|---|---|---|---|---|---|---|---|
| airplane | 96.87 | 96.46 | 93.88 | 96.06 | 96.06 | 96.06 | 96.06 |
|  | **98.36** | 98.07 | 96.65 | 97.48 | 97.48 | 97.48 | 97.48 |
| automobile | 99.00 | 98.90 | 92.73 | 98.81 | 98.81 | 98.81 | 98.81 |
|  | **99.46** | 99.42 | 95.40 | 99.26 | 99.24 | 99.26 | 99.23 |
| bird | 93.40 | 92.44 | 75.86 | 92.39 | 92.39 | 92.39 | 92.39 |
|  | **97.16** | 96.54 | 86.95 | 96.00 | 96.00 | 96.00 | 96.00 |
| cat | 91.40 | 90.51 | 71.11 | 90.43 | 90.43 | 90.43 | 90.43 |
|  | **95.96** | 95.25 | 82.18 | 94.86 | 94.86 | 94.86 | 94.86 |
| deer | 96.65 | 96.13 | 90.06 | 95.77 | 95.77 | 95.77 | 95.77 |
|  | **98.86** | 98.49 | 95.29 | 97.94 | 97.94 | 97.94 | 97.94 |
| dog | 95.14 | 94.41 | 71.72 | 94.12 | 94.12 | 94.12 | 94.12 |
|  | **97.63** | 97.20 | 81.65 | 96.47 | 96.47 | 96.47 | 96.47 |
| frog | 98.23 | 97.85 | 71.35 | 97.61 | 97.61 | 97.61 | 97.61 |
|  | **99.28** | 99.10 | 81.67 | 98.61 | 98.61 | 98.61 | 98.61 |
| horse | 98.29 | 97.95 | 97.28 | 97.69 | 97.69 | 97.69 | 97.69 |
|  | **99.26** | 99.12 | 98.79 | 98.62 | 98.62 | 98.62 | 98.62 |
| ship | 98.35 | 98.15 | 96.92 | 98.01 | 98.01 | 98.01 | 98.01 |
|  | **99.08** | 98.92 | 98.17 | 98.68 | 98.68 | 98.68 | 98.68 |
| truck | 97.33 | 96.92 | 84.05 | 96.63 | 96.63 | 96.63 | 96.63 |
|  | **98.57** | 98.27 | 88.48 | 97.85 | 97.85 | 97.85 | 97.85 |
| Average | 96.47 | 95.97 | 84.50 | 95.75 | 95.75 | 95.75 | 95.75 |
|  | **98.36** | 98.04 | 90.52 | 97.58 | 97.57 | 97.58 | 97.57 |

Table B.8: Average % AUROC over 5 seeds with the HSC loss for each class of the CIFAR-10 dataset. AUROC over unperturbed and perturbed test inputs is shown in the white and gray rows respectively, with the largest value per class emphasized in bold font.

## B.3 MVTecAD

| Label | Fully Trained | CalHead OE | CalHead Spectral | Platt OE | Platt Spectral | $\beta$ OE | $\beta$ Spectral |
|---|---|---|---|---|---|---|---|
| bottle | 67.38 | 55.70 | 56.24 | 72.78 | 72.78 | 72.79 | 72.79 |
| | 65.22 | **100** | 99.02 | **100** | **100** | **100** | **100** |
| cable | 67.76 | 56.07 | 54.29 | 66.03 | 66.03 | 66.03 | 66.03 |
| | 67.93 | 66.36 | 65.39 | **82.99** | **82.99** | **82.99** | **82.99** |
| capsule | 66.90 | 58.88 | 70.95 | 70.58 | 70.58 | 70.58 | 70.58 |
| | 61.50 | 99.82 | **100** | **100** | **100** | **100** | **100** |
| carpet | 37.74 | 47.62 | 49.92 | 43.42 | 43.42 | 43.42 | 43.42 |
| | 39.20 | 99.85 | 99.51 | **100** | **100** | **100** | **100** |
| grid | 55.97 | 57.41 | 38.65 | 60.42 | 60.42 | 60.42 | 60.42 |
| | 57.46 | **100** | 99.95 | **100** | **100** | 99.28 | **100** |
| hazelnut | 64.46 | 58.44 | 56.29 | 68.38 | 68.38 | 68.38 | 68.38 |
| | 64.33 | **99.96** | **99.96** | 98.35 | 98.35 | 98.35 | 98.35 |
| leather | 66.24 | 48.85 | 41.45 | 71.53 | 71.53 | 71.53 | 71.53 |
| | 67.70 | 97.82 | **99.85** | 97.94 | 98.74 | 98.74 | 98.74 |
| metal nut | 50.96 | 51.56 | 57.17 | 49.28 | 49.28 | 49.28 | 49.28 |
| | 51.28 | **95.72** | 89.56 | 94.69 | 94.69 | 94.69 | 94.69 |
| pill | 65.76 | 53.57 | 59.29 | 61.91 | 61.91 | 61.91 | 61.91 |
| | 63.84 | **100** | 99.23 | **100** | **100** | **100** | **100** |
| screw | 27.54 | 31.52 | 32.66 | 27.83 | 27.83 | 27.83 | 27.83 |
| | 30.73 | 72.44 | 71.33 | **94.27** | **94.27** | **94.27** | **94.27** |
| tile | 71.88 | 63.25 | 50.42 | 69.15 | 69.15 | 69.14 | 69.14 |
| | 70.50 | **100** | **100** | **100** | **100** | **100** | **100** |
| toothbrush | 85.83 | 39.22 | 65.67 | 86.22 | 86.22 | 86.22 | 86.22 |
| | 84.83 | 95.33 | 96.00 | **99.94** | **99.94** | **99.94** | **99.94** |
| transistor | 81.98 | 41.57 | 42.79 | 81.89 | 81.89 | 81.89 | 81.89 |
| | 81.12 | 66.48 | 68.60 | **96.08** | **96.08** | **96.08** | **96.08** |
| wood | 88.11 | 80.70 | 83.19 | 87.68 | 87.68 | 87.68 | 87.68 |
| | 63.67 | 99.82 | **100** | **100** | **100** | **100** | **100** |
| zipper | 74.85 | 50.59 | 51.19 | 72.43 | 72.43 | 72.43 | 72.43 |
| | 70.84 | 98.75 | 98.60 | **100** | **100** | 99.98 | **100** |
| Average | 64.89 | 53.00 | 54.01 | 65.97 | 65.97 | 65.97 | 65.97 |
| | 62.68 | 92.82 | 92.47 | 97.62 | **97.67** | 97.62 | **97.67** |

Table B.9: Average % AUROC over 5 seeds with the SVDD loss for each class of the MVTecAD dataset. AUROC over unperturbed and perturbed test inputs is shown in the white and gray rows respectively, with the largest value per class emphasized in bold font.

| Label | Fully Trained | CalHead OE | CalHead Spectral | Platt OE | Platt Spectral | $\beta$ OE | $\beta$ Spectral |
|---|---|---|---|---|---|---|---|
| bottle | 82.79 | 71.11 | 69.35 | 83.25 | 83.25 | 83.25 | 83.25 |
| | 82.60 | 92.63 | 92.78 | **95.25** | **95.25** | **95.25** | **95.25** |
| cable | 85.51 | 63.99 | 59.05 | 87.04 | 87.04 | 87.04 | 87.04 |
| | 85.47 | 76.83 | 74.56 | 94.69 | **94.96** | 94.96 | **94.96** |
| capsule | 60.57 | 52.91 | 38.76 | 67.34 | 67.34 | 67.34 | 67.34 |
| | 60.22 | 81.67 | 57.06 | **87.17** | **87.17** | **87.17** | **87.17** |
| carpet | 60.60 | 48.53 | 51.27 | 60.75 | 60.75 | 60.75 | 60.75 |
| | 60.43 | 66.40 | **72.40** | 64.33 | 64.01 | 64.44 | 64.12 |
| grid | 28.97 | 46.62 | 46.08 | 60.95 | 60.95 | 60.95 | 60.95 |
| | 30.36 | 54.64 | 54.07 | 69.34 | **69.89** | 69.74 | **69.89** |
| hazelnut | 49.04 | 50.86 | 52.29 | 69.64 | 69.64 | 69.64 | 69.64 |
| | 51.60 | 66.86 | 70.51 | **94.75** | **94.75** | 94.74 | **94.75** |
| leather | 54.59 | 47.11 | 42.20 | 38.57 | 38.57 | 38.57 | 38.57 |
| | 52.04 | 54.69 | 50.95 | **68.65** | **68.65** | **68.65** | **68.65** |
| metal nut | 72.40 | 58.12 | 63.57 | 77.48 | 77.48 | 77.48 | 77.48 |
| | 71.21 | 65.53 | 70.51 | **89.43** | **89.43** | **89.43** | **89.43** |
| pill | 79.99 | 62.58 | 65.17 | 79.05 | 79.04 | 79.05 | 79.04 |
| | 79.55 | 99.98 | **99.99** | 93.59 | 93.58 | 93.59 | 93.58 |
| screw | 12.45 | 39.88 | 27.33 | 11.88 | 11.88 | 11.88 | 11.88 |
| | 17.26 | 47.79 | 34.39 | 65.88 | **69.16** | 68.55 | **69.16** |
| tile | 49.28 | 42.55 | 43.56 | 41.25 | 41.25 | 41.25 | 41.25 |
| | 49.83 | **84.91** | 82.56 | 68.51 | 68.51 | 68.51 | 68.51 |
| toothbrush | 69.67 | 72.94 | 70.11 | 73.17 | 73.17 | 73.17 | 73.17 |
| | 69.28 | 82.50 | 80.44 | **83.28** | **83.28** | **83.28** | **83.28** |
| transistor | 70.23 | 67.37 | 60.51 | 87.22 | 87.22 | 87.22 | 87.22 |
| | 69.46 | 78.03 | 72.87 | **97.22** | **97.22** | **97.22** | **97.22** |
| wood | 63.39 | 61.04 | 50.37 | 65.96 | 65.96 | 65.96 | 65.96 |
| | 65.35 | **94.46** | 93.25 | 81.63 | 81.63 | 81.63 | 81.63 |
| zipper | 30.40 | 66.38 | 62.11 | 41.90 | 41.90 | 41.90 | 41.90 |
| | 30.34 | 77.87 | 77.90 | **86.93** | **86.93** | 86.93 | **86.93** |
| Average | 57.99 | 56.80 | 53.45 | 63.03 | 63.03 | 63.03 | 63.03 |
| | 58.33 | 74.99 | 72.28 | 82.71 | 82.96 | 82.94 | **82.97** |

Table B.10: Average % AUROC over 5 seeds with the SSIM loss for each class of the MVTecAD dataset. AUROC over unperturbed and perturbed test inputs is shown in the white and gray rows respectively, with the largest value per class emphasized in bold font.

| Label | Fully Trained | CalHead OE | CalHead Spectral | Platt OE | Platt Spectral | $\beta$ OE | $\beta$ Spectral |
|---|---|---|---|---|---|---|---|
| bottle | 60.67 | 73.37 | 52.44 | 70.52 | 70.52 | 54.51 | 54.51 |
| | 86.83 | **94.14** | 85.16 | 91.33 | 93.62 | 57.08 | 57.08 |
| cable | 63.34 | 58.82 | 60.07 | 59.07 | 59.07 | 53.59 | 53.59 |
| | 86.05 | **88.97** | 75.57 | 83.26 | 83.26 | 53.59 | 53.59 |
| capsule | 55.95 | 59.74 | 54.13 | 58.69 | 58.69 | 50.28 | 50.28 |
| | 92.23 | **96.48** | 88.19 | 96.10 | 96.10 | 50.28 | 50.28 |
| carpet | 67.19 | 49.81 | 83.56 | 49.25 | 49.25 | 49.56 | 49.56 |
| | 86.60 | 86.06 | **95.82** | 81.95 | 83.77 | 50.69 | 50.69 |
| grid | 49.86 | 52.87 | 11.48 | 55.09 | 55.09 | 54.32 | 54.32 |
| | 64.70 | 76.59 | 19.55 | 79.47 | **79.83** | 56.59 | 56.59 |
| hazelnut | 73.81 | 60.91 | 95.14 | 75.19 | 75.19 | 71.86 | 71.86 |
| | 91.41 | 84.44 | **99.74** | 91.55 | 92.68 | 71.86 | 71.86 |
| leather | 66.39 | 69.22 | 87.83 | 69.25 | 69.25 | 56.59 | 56.59 |
| | 85.12 | 85.58 | **97.59** | 85.29 | 85.29 | 58.89 | 58.89 |
| metal nut | 56.62 | 56.02 | 56.59 | 55.31 | 55.31 | 54.99 | 54.99 |
| | 75.93 | **78.80** | 78.40 | 73.29 | 77.39 | 55.11 | 55.11 |
| pill | 70.45 | 70.82 | 66.65 | 73.69 | 73.69 | 54.65 | 54.65 |
| | 94.94 | **98.82** | 88.58 | 97.32 | 97.99 | 56.92 | 56.92 |
| screw | 5.83 | 16.28 | 44.01 | 15.27 | 13.47 | 42.77 | 37.89 |
| | 21.64 | 55.63 | **72.32** | 49.80 | 48.70 | 46.45 | 43.31 |
| tile | 90.96 | 93.52 | 74.09 | 93.40 | 93.40 | 81.59 | 81.59 |
| | 96.82 | 99.01 | 83.46 | 98.56 | **99.10** | 82.20 | 82.20 |
| toothbrush | 72.56 | 70.61 | 54.11 | 69.56 | 69.56 | 51.33 | 51.33 |
| | 94.50 | **95.22** | 76.67 | 94.89 | 95.06 | 51.33 | 51.33 |
| transistor | 62.18 | 48.48 | 31.84 | 49.05 | 49.05 | 53.50 | 53.50 |
| | **87.85** | 82.12 | 68.80 | 78.29 | 81.58 | 53.50 | 53.50 |
| wood | 83.93 | 90.16 | 82.02 | 91.70 | 91.70 | 64.83 | 64.83 |
| | 90.07 | 94.54 | 87.86 | 95.30 | **95.37** | 64.83 | 64.83 |
| zipper | 78.71 | 69.13 | 43.97 | 69.41 | 69.41 | 50.25 | 50.25 |
| | **95.74** | 94.40 | 67.79 | 93.17 | 95.39 | 50.25 | 50.25 |
| Average | 63.90 | 62.65 | 59.86 | 63.63 | 63.51 | 56.31 | 55.98 |
| | 83.36 | **87.39** | 79.03 | 85.97 | 87.01 | 57.30 | 57.10 |

Table B.11: Average % AUROC over 5 seeds with the logistic loss for each class of the MVTecAD dataset. AUROC over unperturbed and perturbed test inputs is shown in the white and gray rows respectively, with the largest value per class emphasized in bold font.

| Label | Fully Trained | CalHead OE | CalHead Spectral | Platt OE | Platt Spectral | $\beta$ OE | $\beta$ Spectral |
|---|---|---|---|---|---|---|---|
| bottle | 74.30 | 66.83 | 58.14 | 72.92 | 72.92 | 72.92 | 72.92 |
|  | 82.82 | **96.37** | 93.90 | 90.37 | 90.45 | 90.45 | 90.45 |
| cable | 58.15 | 54.04 | 51.96 | 59.38 | 59.38 | 59.38 | 59.38 |
|  | 64.29 | **94.87** | 79.47 | 64.74 | 64.75 | 63.85 | 64.75 |
| capsule | 51.88 | 51.11 | 74.08 | 55.34 | 55.34 | 55.34 | 55.34 |
|  | 52.29 | **100** | 97.30 | 59.71 | 60.52 | 60.52 | 60.52 |
| carpet | 63.13 | 66.60 | 36.22 | 54.33 | 54.33 | 54.33 | 54.33 |
|  | 67.78 | **97.47** | 69.45 | 67.02 | 67.02 | 67.02 | 67.02 |
| grid | 53.44 | 43.96 | 92.65 | 53.60 | 53.60 | 53.60 | 53.60 |
|  | 57.63 | 91.13 | **97.16** | 57.07 | 57.46 | 57.46 | 57.46 |
| hazelnut | 63.64 | 59.97 | 39.40 | 69.98 | 69.98 | 69.98 | 69.98 |
|  | 69.51 | **97.25** | 71.42 | 76.95 | 81.62 | 81.62 | 81.62 |
| leather | 74.10 | 78.50 | 29.47 | 76.06 | 76.06 | 76.06 | 76.06 |
|  | 80.19 | **99.86** | 67.86 | 88.01 | 88.08 | 88.08 | 88.08 |
| metal nut | 56.37 | 53.11 | 54.97 | 60.72 | 60.72 | 60.72 | 60.72 |
|  | 73.76 | **97.54** | 85.17 | 77.31 | 78.46 | 78.46 | 78.46 |
| pill | 64.32 | 51.39 | 51.96 | 62.66 | 62.66 | 62.66 | 62.66 |
|  | 72.90 | **98.59** | 88.36 | 94.91 | 96.84 | 96.84 | 96.84 |
| screw | 17.38 | 38.51 | 46.40 | 13.41 | 13.41 | 13.41 | 13.41 |
|  | 33.80 | **72.75** | 61.41 | 27.57 | 26.32 | 28.29 | 28.29 |
| tile | 92.43 | 92.91 | 54.23 | 93.54 | 93.54 | 93.54 | 93.54 |
|  | 92.86 | **99.91** | 74.27 | 93.69 | 93.81 | 93.81 | 93.81 |
| toothbrush | 83.61 | 61.50 | 53.28 | 71.22 | 71.22 | 71.22 | 71.22 |
|  | **100** | 98.28 | 85.83 | 98.17 | 98.50 | 98.50 | 98.50 |
| transistor | 59.27 | 54.44 | 55.86 | 60.81 | 60.81 | 60.81 | 60.81 |
|  | 74.38 | **93.48** | 82.33 | 78.89 | 79.29 | 79.29 | 79.29 |
| wood | 75.50 | 63.76 | 60.49 | 81.11 | 81.11 | 81.11 | 81.11 |
|  | 75.67 | **100** | 83.47 | 81.27 | 81.54 | 81.54 | 81.54 |
| zipper | 63.84 | 57.49 | 61.57 | 56.75 | 56.75 | 56.75 | 56.75 |
|  | 69.16 | **100** | 90.58 | 72.50 | 76.96 | 76.96 | 76.96 |
| Average | 63.42 | 59.61 | 54.71 | 62.79 | 62.79 | 62.79 | 62.79 |
|  | 71.14 | **95.83** | 81.87 | 75.21 | 76.11 | 76.18 | 76.24 |

Table B.12: Average % AUROC over 5 seeds with the HSC loss for each class of the MVTecAD dataset. AUROC over unperturbed and perturbed test inputs is shown in the white and gray rows respectively, with the largest value per class emphasized in bold font.

## B.4 MPDD

| Label | Fully Trained | CalHead OE | CalHead Spectral | Platt OE | Platt Spectral | $\beta$ OE | $\beta$ Spectral |
|---|---|---|---|---|---|---|---|
| bracket black | 62.47 | 56.53 | 47.97 | 58.51 | 58.52 | 58.51 | 58.51 |
| | 63.12 | 74.92 | 78.96 | **80.47** | **80.47** | **80.47** | **80.47** |
| bracket brown | 75.90 | 56.23 | 66.53 | 75.57 | 75.57 | 75.57 | 75.57 |
| | 75.40 | 78.94 | 87.74 | **96.20** | **96.20** | **96.20** | **96.20** |
| bracket white | 66.71 | 47.87 | 47.73 | 63.87 | 63.87 | 63.87 | 63.87 |
| | 65.87 | 64.13 | 80.93 | **88.29** | **88.29** | **88.29** | **88.29** |
| connector | 64.71 | 33.10 | 34.43 | 97.71 | 97.71 | 97.71 | 97.71 |
| | 64.14 | 44.48 | 46.76 | 99.48 | **99.62** | 99.43 | 99.38 |
| metal plate | **100** | 92.00 | 91.11 | 98.96 | 98.96 | 98.96 | 98.96 |
| | **100** | 98.39 | 98.82 | 99.87 | 99.87 | 99.87 | 99.87 |
| tubes | 44.30 | 24.75 | 76.37 | 41.01 | 41.01 | 41.01 | 41.01 |
| | 44.54 | 45.19 | **83.91** | 70.55 | 70.55 | 70.55 | 70.55 |
| Average | 69.02 | 51.75 | 60.69 | 72.61 | 72.61 | 72.61 | 72.61 |
| | 68.85 | 67.67 | 79.52 | 89.14 | **89.17** | 89.13 | 89.13 |

Table B.13: Average % AUROC over 5 seeds with the SVDD loss for each class of the MPDD dataset. AUROC over unperturbed and perturbed test inputs is shown in the white and gray rows respectively, with the largest value per class emphasized in bold font.

| Label | Fully Trained | CalHead OE | CalHead Spectral | Platt OE | Platt Spectral | $\beta$ OE | $\beta$ Spectral |
|---|---|---|---|---|---|---|---|
| bracket black | 41.53 | 75.25 | 63.94 | 43.30 | 43.30 | 43.30 | 43.30 |
| | 41.81 | **84.20** | 77.17 | 57.23 | 57.10 | 57.23 | 57.23 |
| bracket brown | 71.34 | 41.48 | 34.42 | 85.13 | 85.13 | 85.13 | 85.13 |
| | 66.94 | 80.48 | 74.83 | **99.28** | **99.28** | **99.28** | **99.28** |
| bracket white | 71.07 | 57.51 | 62.53 | 71.16 | 71.16 | 71.16 | 71.16 |
| | 70.67 | 59.69 | 63.91 | **87.71** | **87.71** | **87.71** | **87.71** |
| connector | 99.24 | 98.05 | 98.43 | 99.95 | 99.95 | 99.95 | 99.95 |
| | 99.24 | 99.24 | 98.95 | **100** | **100** | **100** | **100** |
| metal plate | 83.02 | 44.05 | 39.47 | 84.78 | 84.78 | 84.78 | 84.78 |
| | 82.38 | 47.15 | 42.68 | **92.75** | 91.79 | **92.75** | **92.75** |
| tubes | 41.02 | 53.43 | 44.12 | 45.91 | 45.91 | 45.91 | 45.91 |
| | 40.54 | **68.92** | 56.33 | 68.59 | 68.59 | 68.59 | 68.59 |
| Average | 67.87 | 61.63 | 57.15 | 71.70 | 71.70 | 71.70 | 71.70 |
| | 66.93 | 73.28 | 68.98 | **84.26** | 84.08 | **84.26** | **84.26** |

Table B.14: Average % AUROC over 5 seeds with the SSIM loss for each class of the MPDD dataset. AUROC over unperturbed and perturbed test inputs is shown in the white and gray rows respectively, with the largest value per class emphasized in bold font.

| Label | Fully Trained | CalHead OE | CalHead Spectral | Platt OE | Platt Spectral | $\beta$ OE | $\beta$ Spectral |
|---|---|---|---|---|---|---|---|
| bracket black | 28.24 | 32.27 | 32.13 | 34.41 | 34.41 | 47.72 | 47.72 |
| | 47.09 | 43.47 | 43.07 | 44.10 | 44.60 | **49.04** | **49.04** |
| bracket brown | 52.68 | 52.76 | 40.95 | 54.37 | 54.37 | 49.62 | 49.62 |
| | 64.83 | **73.36** | 58.57 | 72.78 | 72.78 | 49.62 | 49.62 |
| bracket white | 48.27 | 47.07 | 34.87 | 46.60 | 46.60 | 51.31 | 51.31 |
| | 60.91 | **63.42** | 50.29 | 58.64 | 63.22 | 52.38 | 52.38 |
| connector | 88.19 | 81.43 | 87.48 | 78.81 | 78.81 | 61.81 | 61.81 |
| | **94.90** | 87.29 | 89.71 | 83.76 | 84.81 | 63.57 | 63.57 |
| metal plate | 86.05 | 92.57 | 95.94 | 91.87 | 91.87 | 89.83 | 89.83 |
| | 91.73 | 96.61 | **97.79** | 96.00 | 96.00 | 91.71 | 91.71 |
| tubes | 47.51 | 45.07 | 42.31 | 43.65 | 43.65 | 54.70 | 54.70 |
| | **66.78** | 66.34 | 56.70 | 62.23 | 63.21 | 55.40 | 55.40 |
| Average | 58.49 | 58.53 | 55.61 | 58.29 | 58.29 | 59.16 | 59.16 |
| | 71.04 | **71.75** | 66.02 | 69.58 | 70.77 | 60.28 | 60.28 |

Table B.15: Average % AUROC over 5 seeds with the logistic loss for each class of the MPDD dataset. AUROC over unperturbed and perturbed test inputs is shown in the white and gray rows respectively, with the largest value per class emphasized in bold font.

| Label | Fully Trained | CalHead OE | CalHead Spectral | Platt OE | Platt Spectral | $\beta$ OE | $\beta$ Spectral |
|---|---|---|---|---|---|---|---|
| bracket black | 51.00 | 50.30 | 41.89 | 49.11 | 49.11 | 49.11 | 49.11 |
| | 52.80 | **96.30** | 84.61 | 52.22 | 54.36 | 54.36 | 54.36 |
| bracket brown | 55.77 | 56.41 | 47.15 | 64.58 | 64.58 | 64.58 | 64.58 |
| | 62.90 | **99.49** | 67.84 | 80.84 | 82.55 | 82.55 | 82.55 |
| bracket white | 61.41 | 47.18 | 42.76 | 58.61 | 58.61 | 58.61 | 58.61 |
| | 88.59 | **98.98** | 65.91 | 80.94 | 82.10 | 82.10 | 82.10 |
| connector | 98.19 | 74.90 | 96.10 | 99.71 | 99.71 | 99.71 | 99.71 |
| | 99.90 | 90.38 | 98.62 | **99.95** | **99.95** | **99.95** | **99.95** |
| metal plate | 96.28 | 90.48 | 93.62 | 96.55 | 96.55 | 96.55 | 96.55 |
| | **99.61** | 99.45 | 98.09 | 99.52 | 99.13 | 99.33 | 99.52 |
| tubes | 50.28 | 60.56 | 56.49 | 58.68 | 58.68 | 58.68 | 58.68 |
| | 56.39 | **92.55** | 75.95 | 66.68 | 70.37 | 70.00 | 70.37 |
| Average | 68.82 | 63.30 | 63.00 | 71.21 | 71.21 | 71.21 | 71.21 |
| | 76.70 | **96.19** | 81.84 | 80.03 | 81.41 | 81.38 | 81.48 |

Table B.16: Average % AUROC over 5 seeds with the HSC loss for each class of the MPDD dataset. AUROC over unperturbed and perturbed test inputs is shown in the white and gray rows respectively, with the largest value per class emphasized in bold font.

# Appendix C

# Per-Class Localization Results

## C.1 MPDD

| Label | Fully Trained | Platt OE | Platt Spectral | $\beta$ OE | $\beta$ Spectral |
|---|---|---|---|---|---|
| bracket black | 66.52 | 44.39 | 39.93 | 50.07 | 49.66 |
| | **69.99** | 44.39 | 39.93 | 50.07 | 49.66 |
| bracket brown | 86.93 | 50.39 | 49.09 | 49.54 | 50.49 |
| | **90.82** | 50.39 | 49.09 | 49.54 | 50.49 |
| bracket white | 80.91 | 53.84 | 51.33 | 50.24 | 51.23 |
| | **85.92** | 53.87 | 51.35 | 50.26 | 51.26 |
| connector | 86.12 | 52.62 | 48.27 | 49.96 | 50.08 |
| | **88.13** | 52.79 | 48.40 | 50.11 | 50.20 |
| metal plate | 86.87 | 85.65 | 85.63 | 85.56 | 85.56 |
| | **87.38** | 86.98 | 86.96 | 86.91 | 86.91 |
| tubes | 77.47 | 56.57 | 56.23 | 54.91 | 55.86 |
| | **79.94** | 56.70 | 57.70 | 55.08 | 56.29 |
| Average | 80.80 | 57.25 | 55.08 | 56.71 | 57.15 |
| | **83.70** | 57.52 | 55.57 | 57.00 | 57.47 |

Table C.1: Average % per-pixel AUROC over 5 seeds with the FCDD loss for each class of the MPDD dataset. AUROC over unperturbed and perturbed test inputs is shown in the white and gray rows respectively, with the largest value per class emphasized in bold font.

| Label | Fully Trained | Platt OE | Platt Spectral | $\beta$ OE | $\beta$ Spectral |
|---|---|---|---|---|---|
| bracket black | 23.59 | 9.21 | 7.05 | 14.76 | 14.74 |
| | **27.46** | 9.22 | 7.06 | 14.77 | 14.74 |
| bracket brown | 67.60 | 15.20 | 14.42 | 14.87 | 15.84 |
| | **74.47** | 15.21 | 14.42 | 14.87 | 15.84 |
| bracket white | 53.26 | 16.87 | 13.77 | 14.73 | 16.57 |
| | **64.00** | 16.91 | 13.80 | 14.76 | 16.60 |
| connector | 61.34 | 15.86 | 12.99 | 14.71 | 15.22 |
| | **65.48** | 16.10 | 13.19 | 14.92 | 15.40 |
| metal plate | **74.41** | 47.55 | 47.51 | 47.36 | 47.35 |
| | 72.66 | 50.33 | 50.32 | 50.23 | 50.23 |
| tubes | 56.29 | 23.29 | 22.60 | 22.44 | 23.63 |
| | **59.54** | 23.66 | 25.21 | 22.95 | 24.53 |
| Average | 56.08 | 21.33 | 19.72 | 21.48 | 22.22 |
| | **60.60** | 21.90 | 20.67 | 22.08 | 22.89 |

Table C.2: Average % AUPRO over 5 seeds with the FCDD loss for each class of the MPDD dataset. AUPRO over unperturbed and perturbed test inputs is shown in the white and gray rows respectively, with the largest value per class emphasized in bold font.

| Label | Fully Trained | Platt OE | Platt Spectral | $\beta$ OE | $\beta$ Spectral |
|---|---|---|---|---|---|
| bracket black | 95.04 | 93.01 | 92.99 | 91.99 | 92.04 |
| | **95.04** | 93.02 | 93.01 | 92.01 | 92.06 |
| bracket brown | 87.73 | 87.90 | 87.51 | 87.12 | 86.86 |
| | 88.42 | **90.03** | 89.75 | 89.40 | 89.17 |
| bracket white | 81.79 | 80.34 | 80.81 | 80.07 | 80.54 |
| | 82.08 | 82.40 | **82.79** | 82.09 | 82.50 |
| connector | 88.94 | 89.10 | 89.45 | 88.62 | 88.98 |
| | 89.04 | 89.52 | **89.85** | 89.07 | 89.41 |
| metal plate | 80.39 | 79.38 | 79.47 | 79.26 | 79.27 |
| | 79.97 | 81.09 | **81.19** | 80.98 | 81.00 |
| tubes | 88.06 | 85.62 | 85.64 | 83.12 | 83.07 |
| | **88.33** | 86.03 | 86.05 | 83.62 | 83.57 |
| Average | 86.99 | 85.89 | 85.98 | 85.03 | 85.13 |
| | **87.15** | 87.02 | 87.11 | 86.20 | 86.28 |

Table C.3: Average % per-pixel AUROC over 5 seeds with the SSIM loss for each class of the MPDD dataset. AUROC over unperturbed and perturbed test inputs is shown in the white and gray rows respectively, with the largest value per class emphasized in bold font.

| Label | Fully Trained | Platt OE | Platt Spectral | $\beta$ OE | $\beta$ Spectral |
|---|---|---|---|---|---|
| bracket black | 81.65 | 76.91 | 76.90 | 78.10 | 78.31 |
|  | **81.66** | 76.96 | 76.95 | 78.16 | 78.37 |
| bracket brown | 76.89 | 78.14 | 78.16 | 76.66 | 76.75 |
|  | 77.51 | **78.98** | 78.93 | 77.56 | 77.58 |
| bracket white | 49.49 | 49.59 | 50.56 | 49.66 | 50.41 |
|  | 50.45 | 51.87 | **52.78** | 51.87 | 52.66 |
| connector | 66.37 | 66.02 | 66.86 | 64.75 | 65.78 |
|  | 66.78 | 66.75 | **67.59** | 65.54 | 66.53 |
| metal plate | 43.89 | 41.72 | 41.84 | 41.62 | 41.61 |
|  | 43.75 | 44.11 | **44.26** | 44.00 | 44.02 |
| tubes | 60.20 | 52.65 | 52.75 | 45.28 | 45.12 |
|  | **61.08** | 53.90 | 53.99 | 46.74 | 46.59 |
| Average | 63.08 | 60.84 | 61.18 | 59.34 | 59.67 |
|  | **63.54** | 62.10 | 62.42 | 60.65 | 60.96 |

Table C.4: Average % AUPRO over 5 seeds with the SSIM loss for each class of the MPDD dataset. AUPRO over unperturbed and perturbed test inputs is shown in the white and gray rows respectively, with the largest value per class emphasized in bold font.

| Label | Fully Trained | Platt OE | Platt Spectral | $\beta$ OE | $\beta$ Spectral |
|---|---|---|---|---|---|
| bracket black | 31.74 | 36.38 | 38.56 | 50.81 | 49.80 |
|  | 34.38 | 39.84 | 41.93 | **51.82** | 50.75 |
| bracket brown | 33.21 | 35.02 | 35.40 | 49.27 | 49.21 |
|  | 36.09 | 36.47 | 36.84 | **50.08** | 50.03 |
| bracket white | 9.81 | 19.36 | 20.29 | 42.92 | 44.02 |
|  | 11.64 | 21.37 | 22.18 | 44.24 | **45.32** |
| connector | 36.61 | 34.13 | 34.23 | 46.10 | 45.93 |
|  | 39.39 | 36.22 | 36.32 | **47.23** | 47.05 |
| metal plate | 80.06 | 76.84 | 76.85 | 75.88 | 75.90 |
|  | **80.32** | 77.11 | 77.12 | 76.13 | 76.15 |
| tubes | 67.12 | 64.91 | 65.59 | 64.92 | 65.21 |
|  | **70.35** | 68.50 | 69.06 | 68.40 | 68.65 |
| Average | 43.09 | 44.44 | 45.15 | 54.98 | 55.01 |
|  | 45.36 | 46.59 | 47.24 | 56.32 | **56.33** |

Table C.5: Average % per-pixel AUROC over 5 seeds with the logistic loss for each class of the MPDD dataset. AUROC over unperturbed and perturbed test inputs is shown in the white and gray rows respectively, with the largest value per class emphasized in bold font.

| Label | Fully Trained | Platt OE | Platt Spectral | $\beta$ OE | $\beta$ Spectral |
|---|---|---|---|---|---|
| bracket black | 10.42 | 13.58 | 15.87 | 18.60 | 18.69 |
| | 12.42 | 16.18 | 18.60 | **19.88** | 19.81 |
| bracket brown | 17.89 | 17.05 | 16.74 | 20.24 | 20.42 |
| | 20.33 | 18.67 | 18.34 | 21.59 | **21.79** |
| bracket white | 0.81 | 2.68 | 3.27 | 9.17 | 9.80 |
| | 1.49 | 3.68 | 4.20 | 10.24 | **10.83** |
| connector | 15.46 | 13.12 | 13.18 | 16.08 | 16.02 |
| | **17.78** | 13.94 | 14.02 | 16.81 | 16.64 |
| metal plate | 33.25 | 30.48 | 30.46 | 27.92 | 27.94 |
| | **33.35** | 30.78 | 30.75 | 28.27 | 28.28 |
| tubes | 36.57 | 33.59 | 34.01 | 33.70 | 33.87 |
| | **40.02** | 37.58 | 37.89 | 37.84 | 37.89 |
| Average | 19.07 | 18.42 | 18.92 | 20.95 | 21.12 |
| | 20.90 | 20.14 | 20.63 | 22.44 | **22.54** |

Table C.6: Average % AUPRO over 5 seeds with the logistic loss for each class of the MPDD dataset. AUPRO over unperturbed and perturbed test inputs is shown in the white and gray rows respectively, with the largest value per class emphasized in bold font.

## C.2 MVTecAD

| Label | Fully Trained | Platt OE | Platt Spectral | $\beta$ OE | $\beta$ Spectral |
|---|---|---|---|---|---|
| bottle | 73.75 | 75.16 | 75.14 | 74.85 | 74.94 |
|  | 75.72 | **77.03** | 77.01 | 76.89 | 76.96 |
| cable | 82.56 | 66.35 | 64.89 | 64.68 | 64.54 |
|  | **84.16** | 66.59 | 65.12 | 64.95 | 64.81 |
| capsule | **60.31** | 55.78 | 57.07 | 54.78 | 55.02 |
|  | 59.66 | 57.41 | 58.74 | 56.55 | 56.78 |
| carpet | 84.06 | 77.21 | 76.59 | 75.79 | 75.15 |
|  | **85.47** | 81.70 | 81.59 | 80.48 | 80.57 |
| grid | 66.83 | 54.45 | 54.50 | 53.94 | 53.82 |
|  | **70.21** | 55.96 | 55.98 | 55.73 | 55.56 |
| hazelnut | 71.15 | 70.42 | 72.70 | 65.81 | 66.31 |
|  | 73.82 | 72.21 | **74.38** | 67.95 | 68.39 |
| leather | 89.76 | 78.27 | 78.24 | 79.25 | 79.67 |
|  | **91.41** | 79.84 | 79.76 | 80.99 | 81.29 |
| metal nut | **90.06** | 70.38 | 70.79 | 70.04 | 70.32 |
|  | 90.03 | 75.80 | 75.52 | 75.34 | 75.07 |
| pill | 69.74 | 65.09 | 65.37 | 62.79 | 61.49 |
|  | **70.34** | 66.59 | 67.25 | 64.27 | 63.42 |
| screw | **81.88** | 44.53 | 45.50 | 49.47 | 49.86 |
|  | 80.66 | 44.75 | 45.69 | 49.70 | 50.09 |
| tile | 95.33 | 79.62 | 80.24 | 80.10 | 80.46 |
|  | **96.00** | 81.60 | 82.18 | 82.00 | 82.38 |
| toothbrush | 83.38 | 63.40 | 63.77 | 62.87 | 62.96 |
|  | **85.36** | 65.02 | 65.38 | 64.52 | 64.60 |
| transistor | 69.28 | 57.06 | 56.61 | 55.57 | 55.71 |
|  | **76.32** | 59.38 | 58.93 | 57.94 | 58.06 |
| wood | 68.58 | 64.88 | 65.99 | 65.47 | 66.11 |
|  | **71.39** | 66.47 | 67.74 | 67.35 | 68.18 |
| zipper | 78.91 | 48.68 | 49.36 | 53.09 | 53.67 |
|  | **81.89** | 50.18 | 50.93 | 54.70 | 55.42 |
| Average | 77.71 | 64.75 | 65.12 | 64.57 | 64.67 |
|  | **79.50** | 66.70 | 67.08 | 66.62 | 66.77 |

Table C.7: Average % per-pixel AUROC over 5 seeds with the FCDD loss for each class of the MVTecAD dataset. AUROC over unperturbed and perturbed test inputs is shown in the white and gray rows respectively, with the largest value per class emphasized in bold font.

| Label | Fully Trained | Platt OE | Platt Spectral | $\beta$ OE | $\beta$ Spectral |
|---|---|---|---|---|---|
| bottle | 60.29 | 52.52 | 52.38 | 51.61 | 51.65 |
| | **62.87** | 55.16 | 55.00 | 54.62 | 54.68 |
| cable | 50.30 | 24.22 | 22.35 | 23.49 | 23.50 |
| | **53.78** | 24.71 | 22.83 | 24.05 | 24.06 |
| capsule | **29.16** | 21.63 | 21.57 | 20.55 | 21.48 |
| | 27.59 | 22.99 | 22.99 | 22.17 | 23.03 |
| carpet | 36.04 | 29.24 | 30.70 | 27.10 | 28.63 |
| | 39.75 | 37.36 | **40.06** | 34.43 | 37.52 |
| grid | 41.84 | 21.58 | 21.61 | 20.19 | 20.02 |
| | **47.21** | 23.85 | 23.73 | 23.14 | 22.88 |
| hazelnut | 57.16 | 57.34 | 57.99 | 55.04 | 55.44 |
| | 58.48 | 58.83 | **59.39** | 56.99 | 57.30 |
| leather | 67.78 | 48.47 | 48.41 | 48.56 | 48.84 |
| | **71.00** | 50.61 | 50.78 | 51.83 | 52.14 |
| metal nut | **52.32** | 30.66 | 31.66 | 30.10 | 31.02 |
| | 51.83 | 35.10 | 36.01 | 34.73 | 35.68 |
| pill | 32.44 | 19.46 | 19.21 | 18.11 | 17.81 |
| | **36.77** | 20.56 | 20.36 | 19.42 | 19.23 |
| screw | **47.32** | 11.61 | 10.10 | 14.67 | 14.71 |
| | 44.69 | 12.27 | 10.79 | 15.40 | 15.43 |
| tile | 87.32 | 53.30 | 53.63 | 53.22 | 53.09 |
| | **88.04** | 56.72 | 56.87 | 56.60 | 56.53 |
| toothbrush | 35.16 | 20.28 | 20.42 | 19.99 | 19.71 |
| | **38.57** | 21.31 | 21.52 | 21.07 | 20.88 |
| transistor | 42.04 | 22.52 | 21.52 | 21.55 | 21.53 |
| | **50.66** | 26.08 | 25.08 | 25.34 | 25.30 |
| wood | 33.94 | 31.31 | 34.74 | 30.43 | 33.56 |
| | **38.46** | 34.65 | 38.19 | 34.65 | 37.84 |
| zipper | 55.46 | 16.88 | 17.56 | 20.77 | 21.35 |
| | **58.60** | 18.79 | 19.69 | 22.95 | 23.77 |
| Average | 48.57 | 30.73 | 30.92 | 30.36 | 30.82 |
| | **51.22** | 33.27 | 33.55 | 33.16 | 33.75 |

Table C.8: Average % AUPRO over 5 seeds with the FCDD loss for each class of the MVTecAD dataset. AUPRO over unperturbed and perturbed test inputs is shown in the white and gray rows respectively, with the largest value per class emphasized in bold font.

| Label | Fully Trained | Platt OE | Platt Spectral | $\beta$ OE | $\beta$ Spectral |
|---|---|---|---|---|---|
| bottle | 46.62 | 45.55 | 45.86 | 45.59 | 45.94 |
| | **47.01** | 46.59 | 46.92 | 46.61 | 46.97 |
| cable | 63.87 | 62.87 | 63.40 | 62.17 | 62.49 |
| | 64.20 | 64.65 | **65.22** | 63.86 | 64.23 |
| capsule | 83.54 | 81.99 | 82.76 | 81.44 | 81.59 |
| | 84.00 | 84.04 | **84.52** | 83.48 | 83.59 |
| carpet | 50.43 | 50.55 | 50.63 | 49.93 | 50.28 |
| | 50.60 | 53.16 | **53.20** | 52.48 | 52.76 |
| grid | 54.58 | 53.94 | 53.82 | 53.87 | 53.76 |
| | **54.88** | 54.87 | 54.77 | 54.73 | 54.65 |
| hazelnut | 96.38 | 96.18 | 96.13 | 96.15 | 96.13 |
| | 96.64 | **96.71** | 96.63 | 96.69 | 96.65 |
| leather | 78.68 | 75.79 | 74.96 | 76.37 | 77.69 |
| | **80.07** | 78.31 | 77.39 | 78.91 | 80.04 |
| metal nut | 66.35 | 65.87 | 66.10 | 66.00 | 66.29 |
| | 66.43 | 67.58 | 67.78 | 67.66 | **67.91** |
| pill | 49.09 | 53.12 | 53.30 | 52.86 | 52.89 |
| | 50.78 | 55.42 | **55.51** | 55.14 | 55.12 |
| screw | 92.17 | 92.00 | 92.31 | 92.02 | 92.10 |
| | 92.69 | 92.73 | **92.94** | 92.70 | 92.76 |
| tile | 51.16 | 51.21 | 51.17 | 50.91 | 50.92 |
| | 50.93 | 53.36 | **53.36** | 53.06 | 53.12 |
| toothbrush | 78.36 | 78.80 | 78.84 | 78.84 | 78.86 |
| | 78.33 | 79.30 | 79.35 | 79.33 | **79.35** |
| transistor | 58.74 | 57.04 | 57.05 | 56.99 | 57.06 |
| | **58.88** | 58.62 | 58.62 | 58.53 | 58.60 |
| wood | 61.04 | 60.45 | 60.55 | 60.43 | 60.54 |
| | 62.09 | 63.85 | **64.02** | 63.76 | 63.94 |
| zipper | 79.41 | 76.65 | 76.68 | 76.83 | 77.12 |
| | **79.49** | 78.70 | 78.66 | 78.73 | 79.07 |
| Average | 67.36 | 66.80 | 66.91 | 66.69 | 66.91 |
| | 67.80 | 68.53 | **68.59** | 68.38 | 68.58 |

Table C.9: Average % per-pixel AUROC over 5 seeds with the SSIM loss for each class of the MVTecAD dataset. AUROC over unperturbed and perturbed test inputs is shown in the white and gray rows respectively, with the largest value per class emphasized in bold font.

| Label | Fully Trained | Platt OE | Platt Spectral | $\beta$ OE | $\beta$ Spectral |
|---|---|---|---|---|---|
| bottle | 4.18 | 4.33 | 4.62 | 4.63 | 5.01 |
| | 4.68 | 4.82 | 5.12 | 5.18 | **5.57** |
| cable | 25.57 | 25.22 | 25.15 | 24.38 | 24.48 |
| | 26.25 | **26.53** | 26.51 | 25.62 | 25.78 |
| capsule | 60.20 | 56.69 | 56.54 | 56.18 | 56.10 |
| | **60.96** | 59.06 | 58.92 | 58.47 | 58.38 |
| carpet | 11.30 | 11.68 | 11.80 | 11.04 | 11.25 |
| | 11.31 | 12.98 | **13.07** | 12.24 | 12.44 |
| grid | 13.98 | 14.91 | 14.96 | 14.87 | 14.93 |
| | 14.12 | 15.42 | **15.46** | 15.37 | 15.40 |
| hazelnut | 87.56 | 86.70 | 86.48 | 86.52 | 86.39 |
| | **88.44** | 88.04 | 87.77 | 87.88 | 87.73 |
| leather | 71.16 | 64.60 | 60.71 | 66.36 | 68.87 |
| | **74.87** | 68.58 | 64.37 | 70.53 | 72.68 |
| metal nut | 23.56 | 23.52 | 23.50 | 23.72 | 23.95 |
| | 24.19 | 25.37 | 25.35 | 25.50 | **25.70** |
| pill | 5.85 | 8.74 | 8.91 | 8.81 | 8.91 |
| | 6.37 | 10.05 | **10.20** | 10.10 | 10.18 |
| screw | 70.84 | 70.30 | 71.20 | 70.31 | 70.69 |
| | 72.54 | 72.53 | **73.27** | 72.40 | 72.72 |
| tile | 15.07 | 14.66 | 14.77 | 14.36 | 14.56 |
| | 15.05 | 16.13 | **16.39** | 15.82 | 16.18 |
| toothbrush | 35.82 | 36.58 | 36.62 | 36.71 | 36.86 |
| | 35.93 | 37.30 | 37.33 | 37.39 | **37.54** |
| transistor | 17.89 | 17.59 | 17.48 | 17.10 | 17.02 |
| | 18.23 | **18.44** | 18.36 | 17.92 | 17.86 |
| wood | 21.46 | 20.75 | 20.73 | 20.62 | 20.76 |
| | 23.23 | 24.02 | 24.04 | 23.85 | **24.08** |
| zipper | 43.42 | 38.08 | 38.50 | 38.34 | 38.89 |
| | **43.53** | 39.95 | 40.38 | 40.16 | 40.74 |
| Average | 33.86 | 32.96 | 32.80 | 32.93 | 33.24 |
| | 34.65 | 34.61 | 34.44 | 34.56 | **34.87** |

Table C.10: Average % AUPRO over 5 seeds with the SSIM loss for each class of the MVTecAD dataset. AUPRO over unperturbed and perturbed test inputs is shown in the white and gray rows respectively, with the largest value per class emphasized in bold font.

| Label | Fully Trained | Platt OE | Platt Spectral | $\beta$ OE | $\beta$ Spectral |
|---|---|---|---|---|---|
| bottle | 55.23 | 61.64 | 60.96 | 56.18 | 56.62 |
| | 57.71 | **63.51** | 63.17 | 57.31 | 57.65 |
| cable | 70.28 | 59.41 | 61.28 | 60.69 | 63.07 |
| | **72.26** | 61.10 | 62.89 | 62.25 | 64.18 |
| capsule | 57.57 | 51.60 | 51.28 | 49.84 | 49.85 |
| | **60.15** | 54.12 | 53.79 | 52.08 | 52.08 |
| carpet | 62.40 | 64.29 | 60.86 | 56.23 | 55.24 |
| | 65.19 | **66.01** | 62.62 | 57.10 | 56.11 |
| grid | 64.97 | 62.35 | 62.32 | 61.43 | 60.93 |
| | **69.31** | 67.79 | 67.61 | 64.13 | 63.76 |
| hazelnut | **70.68** | 65.93 | 63.26 | 62.88 | |
| | 69.29 | 68.21 | 66.99 | 64.52 | 64.14 |
| leather | 74.53 | 74.88 | 74.45 | 69.15 | 68.79 |
| | 77.10 | **77.85** | 77.34 | 72.45 | 71.86 |
| metal nut | 55.61 | 51.38 | 51.75 | 49.19 | 49.79 |
| | **58.67** | 53.56 | 53.90 | 50.43 | 50.95 |
| pill | 70.21 | 67.46 | 66.73 | 62.15 | 61.37 |
| | **71.94** | 70.01 | 69.47 | 64.71 | 64.06 |
| screw | 59.26 | 51.60 | 50.60 | 55.18 | 53.72 |
| | **62.26** | 53.18 | 52.25 | 56.54 | 54.88 |
| tile | 79.09 | 79.78 | 79.76 | 79.05 | 78.97 |
| | 79.94 | **80.92** | 80.90 | 80.20 | 80.13 |
| toothbrush | 35.72 | 41.41 | 41.57 | 43.74 | 43.89 |
| | 37.60 | 42.82 | 42.95 | 45.00 | **45.13** |
| transistor | 78.85 | 71.69 | 71.38 | 69.63 | 69.42 |
| | **80.39** | 73.21 | 72.83 | 70.53 | 70.28 |
| wood | 73.46 | 71.44 | 71.33 | 67.50 | 67.55 |
| | **75.26** | 74.05 | 73.89 | 69.71 | 69.78 |
| zipper | 64.03 | 53.92 | 55.04 | 51.74 | 52.50 |
| | **72.48** | 59.75 | 60.79 | 56.05 | 56.62 |
| Average | 64.79 | 61.92 | 61.59 | 59.66 | 59.64 |
| | **67.30** | 64.41 | 64.09 | 61.53 | 61.44 |

Table C.11: Average % per-pixel AUROC over 5 seeds with the logistic loss for each class of the MVTecAD dataset. AUROC over unperturbed and perturbed test inputs is shown in the white and gray rows respectively, with the largest value per class emphasized in bold font.

| Label | Fully Trained | Platt OE | Platt Spectral | $\beta$ OE | $\beta$ Spectral |
|---|---|---|---|---|---|
| bottle | 16.28 | 18.78 | 18.91 | 17.19 | 17.34 |
|  | 17.82 | 20.49 | **20.77** | 18.02 | 18.29 |
| cable | 24.24 | 19.89 | 21.92 | 18.61 | 20.24 |
|  | **25.82** | 20.92 | 23.40 | 19.66 | 21.43 |
| capsule | 20.05 | 16.88 | 17.02 | 15.26 | 14.97 |
|  | **22.44** | 19.26 | 19.41 | 17.85 | 17.51 |
| carpet | 30.89 | 32.86 | 29.50 | 21.43 | 20.25 |
|  | 34.51 | **35.14** | 31.72 | 22.45 | 21.30 |
| grid | 29.90 | 28.31 | 28.18 | 27.07 | 26.81 |
|  | **33.39** | 32.38 | 32.20 | 30.37 | 30.19 |
| hazelnut | **58.65** | 55.52 | 53.76 | 47.95 | 47.58 |
|  | 57.12 | 57.32 | 55.61 | 49.68 | 49.29 |
| leather | 42.67 | 42.28 | 41.67 | 37.90 | 37.14 |
|  | **45.97** | 45.88 | 45.19 | 41.46 | 40.46 |
| metal nut | 23.77 | 21.36 | 21.27 | 19.29 | 18.99 |
|  | **27.02** | 23.96 | 23.86 | 20.99 | 20.69 |
| pill | 36.97 | 30.50 | 29.19 | 24.35 | 23.21 |
|  | **38.85** | 32.58 | 31.42 | 26.58 | 25.50 |
| screw | 34.22 | 25.73 | 23.84 | 25.14 | 22.63 |
|  | **37.50** | 27.33 | 25.16 | 26.80 | 24.07 |
| tile | 25.82 | 29.38 | 29.33 | 29.40 | 29.34 |
|  | 26.16 | 30.09 | 30.03 | **30.13** | 30.04 |
| toothbrush | 7.11 | 9.03 | 9.07 | 10.16 | 10.17 |
|  | 7.93 | 9.92 | 9.93 | **10.83** | 10.81 |
| transistor | 23.83 | 23.50 | 23.34 | 21.84 | 21.36 |
|  | **25.66** | 25.40 | 25.23 | 22.76 | 22.25 |
| wood | 35.86 | 34.21 | 33.90 | 29.50 | 29.50 |
|  | 36.36 | **36.37** | 36.02 | 32.03 | 32.04 |
| zipper | 28.67 | 19.76 | 20.77 | 18.50 | 19.21 |
|  | **36.40** | 22.96 | 25.43 | 23.28 | 24.40 |
| Average | 29.26 | 27.20 | 26.78 | 24.24 | 23.92 |
|  | **31.53** | 29.33 | 29.03 | 26.19 | 25.88 |

Table C.12: Average % AUPRO over 5 seeds with the logistic loss for each class of the MVTecAD dataset. AUPRO over unperturbed and perturbed test inputs is shown in the white and gray rows respectively, with the largest value per class emphasized in bold font.